\documentclass[10pt,twocolumn,letterpaper]{article}

\usepackage[pagenumbers]{cvpr} %

\usepackage[dvipsnames]{xcolor}

\usepackage{amsmath}

\usepackage{comment}
\definecolor{cvprblue}{rgb}{0.21,0.49,0.74}
\usepackage[pagebackref,breaklinks,colorlinks,citecolor=cvprblue]{hyperref}

\def\paperID{34} %
\def\confName{EarthVision}
\def\confYear{2024}

\title{SyntStereo2Real: Edge-Aware GAN for Remote Sensing Image-to-Image Translation while Maintaining Stereo Constraint}
\newcounter{daggerfootnote}

\author{Vasudha Venkatesan$^{1}$\thanks{Work done during the completion of master's thesis at DLR}
\and
Daniel Panangian$^2$ 
\and
Mario {Fuentes Reyes}$^2$
\and 
Ksenia Bittner$^2$\thanks{Corresponding author}
\\
\and 
$^1$University of Freiburg\\
\tt\small{venkatev@informatik.uni-freiburg.de}
\and 
$^2$German Aerospace Center (DLR)\\
\tt\small{\{daniel.panangian,mario.fuentesreyes,ksenia.bittner\}@dlr.de}
}

\begin{document}
\maketitle
\begin{abstract}
In the field of remote sensing, the scarcity of stereo-matched and particularly lack of accurate ground truth data often hinders the training of deep neural networks. The use of synthetically generated images as an alternative, alleviates this problem but suffers from the problem of domain generalization. Unifying the capabilities of image-to-image translation and stereo-matching presents an effective solution to address the issue of domain generalization. Current methods involve combining two networks—an unpaired image-to-image translation network and a stereo-matching network—while jointly optimizing them. We propose an edge-aware GAN-based network that effectively tackles both tasks simultaneously. We obtain edge maps of input images from the Sobel operator and use it as an additional input to the encoder in the generator to enforce geometric consistency during translation. We additionally include a warping loss calculated from the translated images to maintain the stereo consistency. We demonstrate that our model produces qualitatively and quantitatively superior results than existing models, and its applicability extends to diverse domains, including autonomous driving.
\end{abstract}
 
\begin{figure}[ht!]
    \begin{subfigure}{\linewidth}
        \begin{subfigure}{.322\linewidth}
            \includegraphics[width=\linewidth]{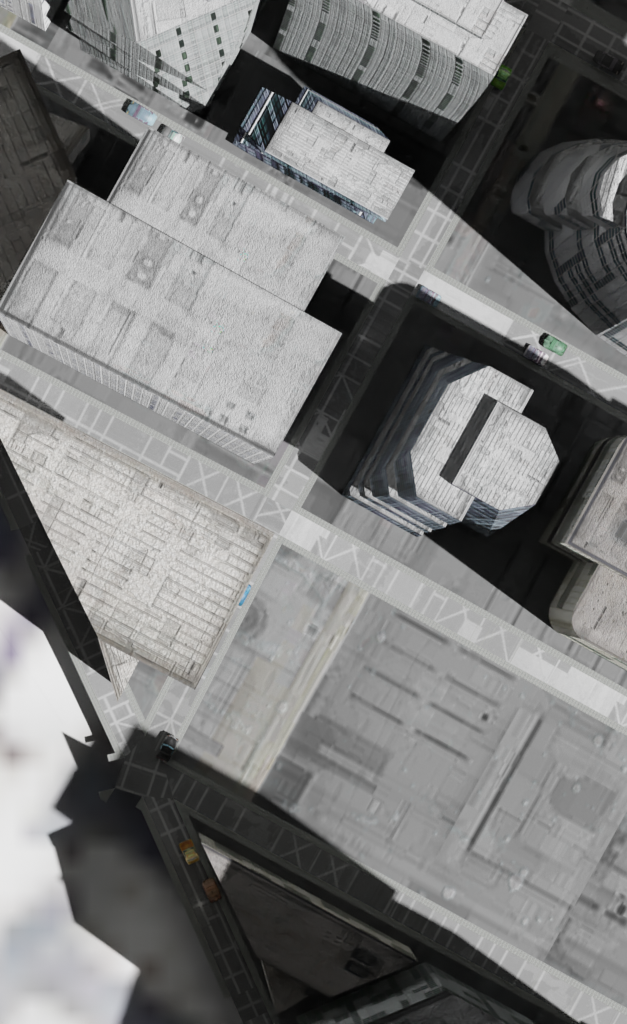}
        \end{subfigure}
        \begin{subfigure}{.322\linewidth}
            \includegraphics[width=\linewidth]{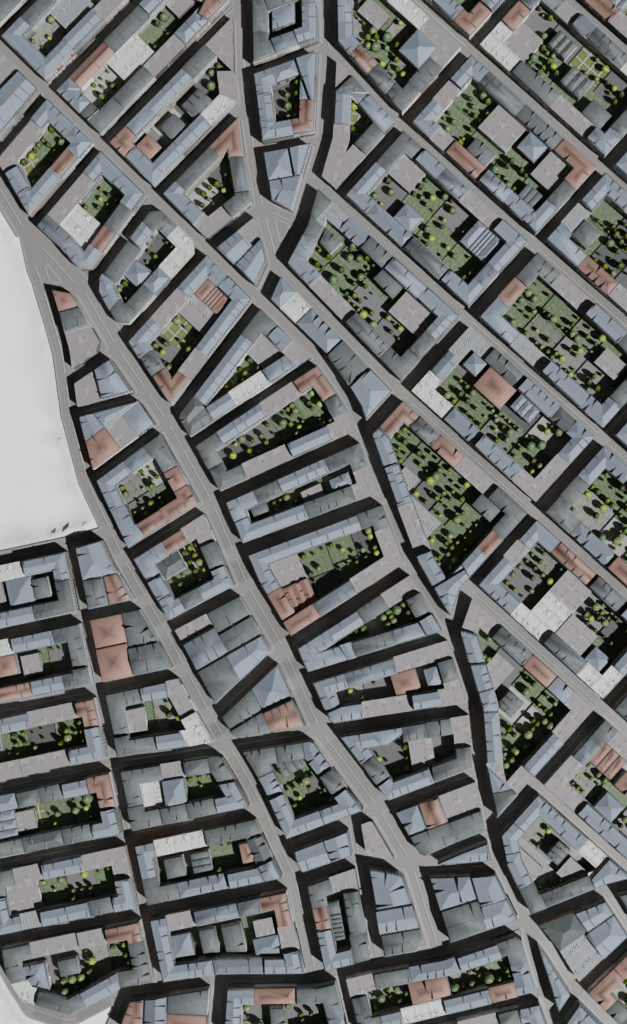}
        \end{subfigure} 
        \begin{subfigure}{.322\linewidth}
            \includegraphics[width=\linewidth]{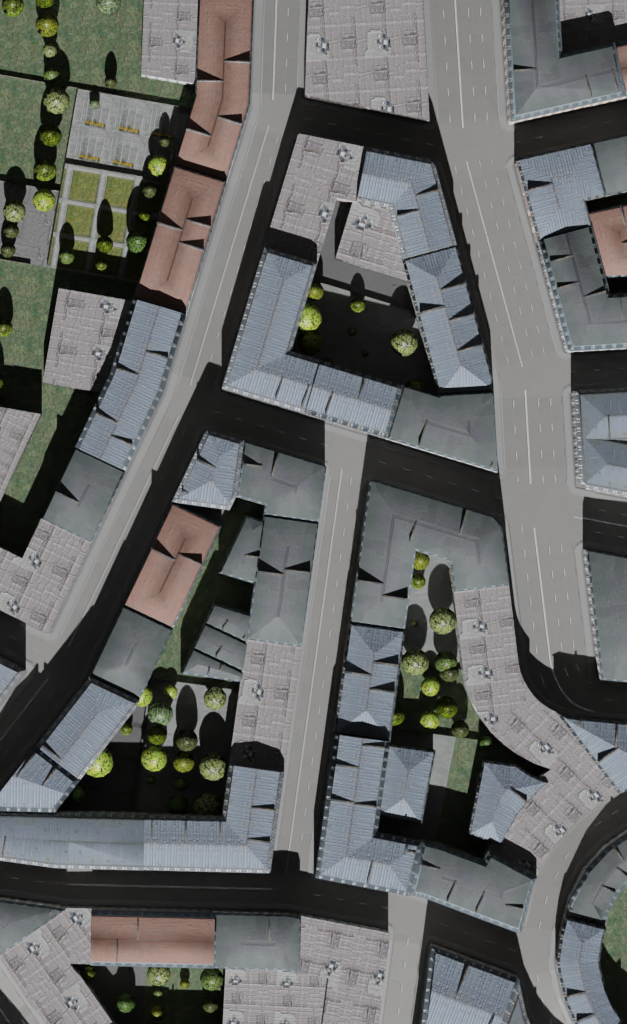}
        \end{subfigure}
        \caption{Original}
    \end{subfigure}
    \begin{subfigure}{\linewidth}
        \begin{subfigure}{.322\linewidth}
            \includegraphics[width=\linewidth]{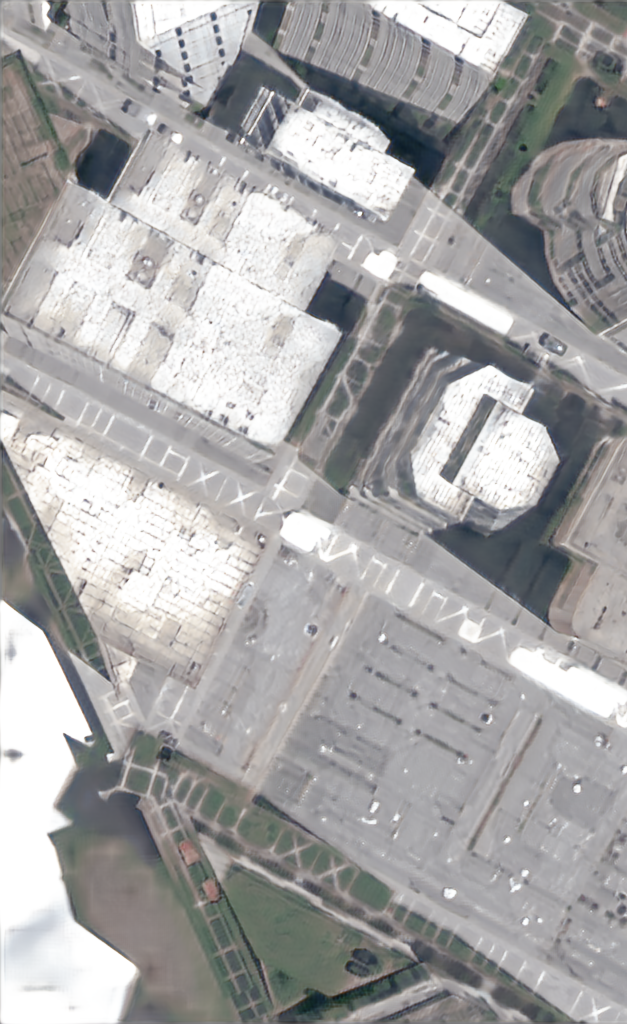}
        \end{subfigure}
        \begin{subfigure}{.322\linewidth}
            \includegraphics[width=\linewidth]{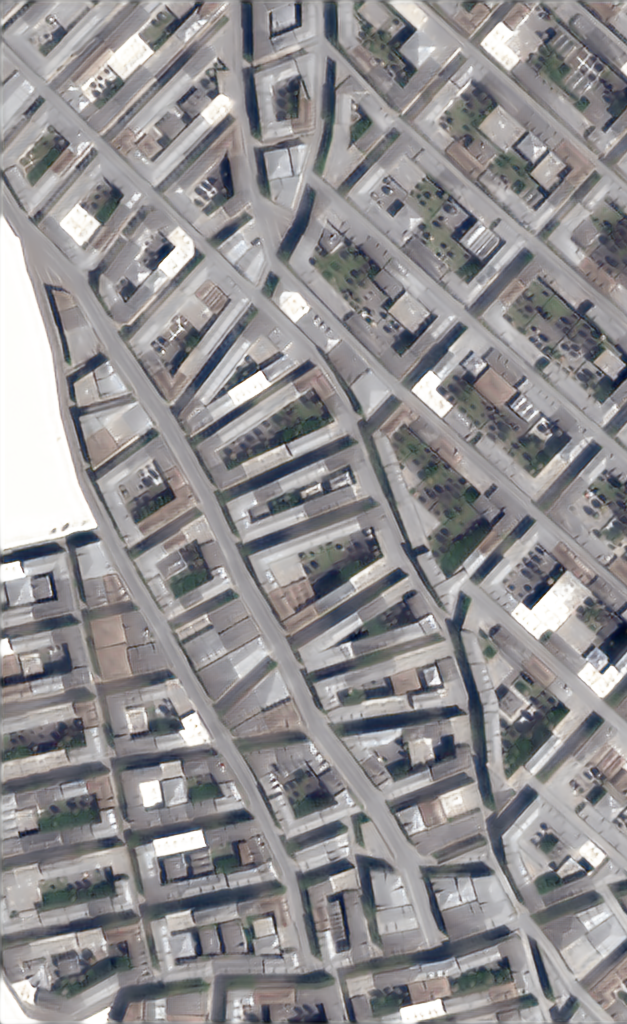}
        \end{subfigure}
        \begin{subfigure}{.322\linewidth}
            \includegraphics[width=\linewidth]{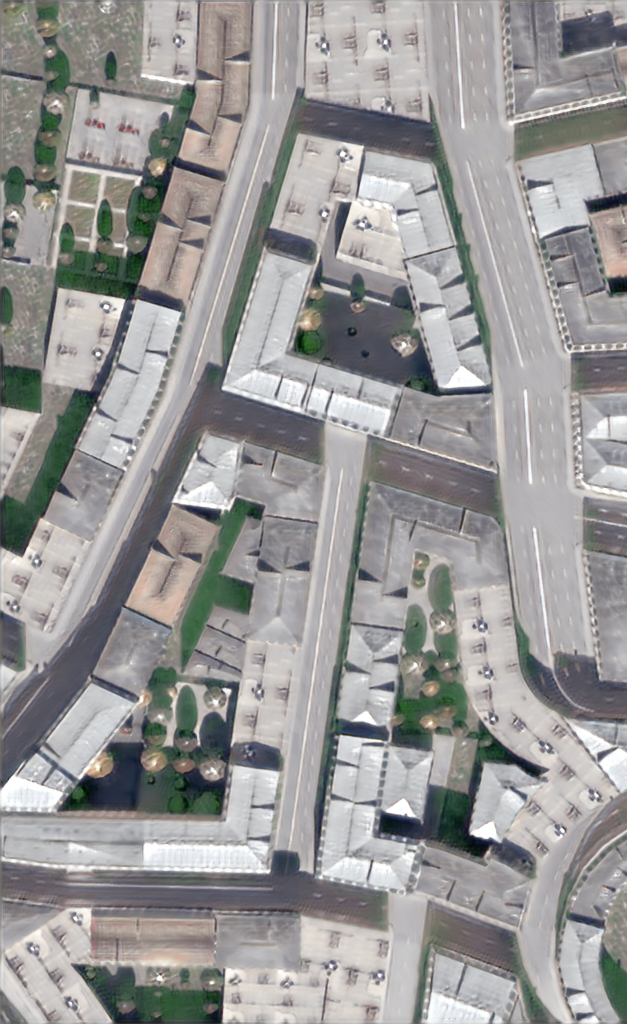}
        \end{subfigure}
        \caption{Translated}
    \end{subfigure} 
    \caption{Examples of aerial scene translated by SyntStereo2Real. Our model can produce semantic-consistent realistic translations.}
\end{figure}

\section{Introduction}
\begin{figure}
    \begin{subfigure}{\linewidth}
    \centering
        \begin{subfigure}{.23\linewidth}
            \includegraphics[width=\linewidth]{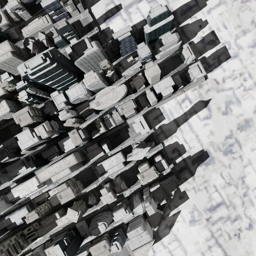}
        \end{subfigure}
        \begin{subfigure}{.23\linewidth}
            \includegraphics[width=\linewidth]{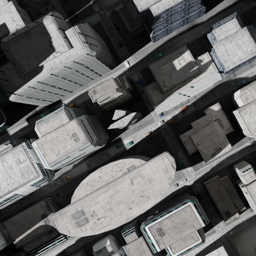}
        \end{subfigure} 
        \begin{subfigure}{.23\linewidth}
            \includegraphics[width=\linewidth]{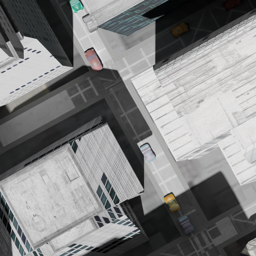}
        \end{subfigure}
            \begin{subfigure}{.23\linewidth}
        \includegraphics[width=\linewidth]{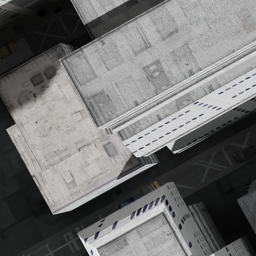}
    \end{subfigure}
        \caption{Original}
    \end{subfigure}
    \begin{subfigure}{\linewidth}
        \centering
        \begin{subfigure}{.23\linewidth}
            \includegraphics[width=\linewidth]{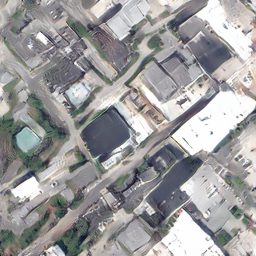}
        \end{subfigure}
        \begin{subfigure}{.23\linewidth}
            \includegraphics[width=\linewidth]{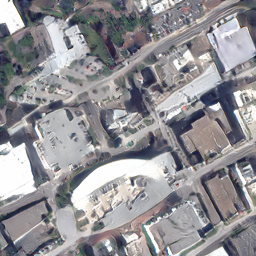}
        \end{subfigure}
        \begin{subfigure}{.23\linewidth}
            \includegraphics[width=\linewidth]{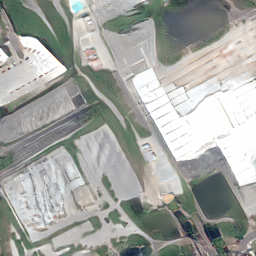}
        \end{subfigure}
        \begin{subfigure}{.23\linewidth}
    \includegraphics[width=\linewidth]{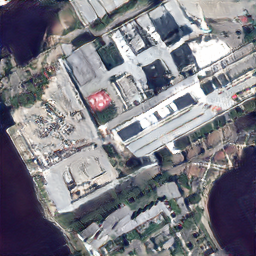}
\end{subfigure}
        \caption{Translated}
    \end{subfigure} 
    \caption{Aerial images translated using CUT \cite{CUT}. The model tends to hallucinate when translating images with diverse scenes, where the target distribution is more likely to be unbalanced. }
    \label{fig:Figure2}
\end{figure}

The challenges in obtaining ground truth images in the remote sensing domain stem from the difficulty in capturing matching images due to temporal changes, sparse measurements and a significantly large baseline. The correspondence tasks like disparity estimation or stereo reconstruction for these images, can be both cumbersome and expensive. The concept of using synthetic data for training deep neural networks arises from the persistent challenges posed by data scarcity, privacy concerns, and the overall difficulty in acquiring authentic data. Synthetic data provides essential ground truth such as accurate labels and stereo disparity maps information for training machine learning models. While the synthetic data is obtained from a simulation of real-world scenario, it may not perfectly represent the complexities and variations in real-world data. This can result in domain shift, where the model struggles to generalize to real world data. Unpaired image-to-image translation algorithms have been used to address the problem of domain shift~\cite{unit,munit,CUT}. They provide promising results to reduce the domain gap between the domains. However they can alter the structural information of the image as shown in \cref{fig:Figure2}. This can pose as a serious challenge when training for downstream tasks as the translated images do not align with their corresponding labels.

Our approach focuses on the specific task of translating synthetic images to realistic domain while maintaining the stereo constraints, which means that pixels do not move within or to another epipolar line and we particularly address the two-view image case. Some of the existing methods such as StereoGAN~\cite{stereogan} have addressed this task for autonomous driving datasets with joint optimization of image translation and disparity estimation networks. Images from remote sensing domain are rich with diverse content. Existing methods suffer from the problem of an increased likelihood of hallucinations and discrepancies in disparity. 

\begin{figure*}[!t]
      \includegraphics[width=\linewidth]{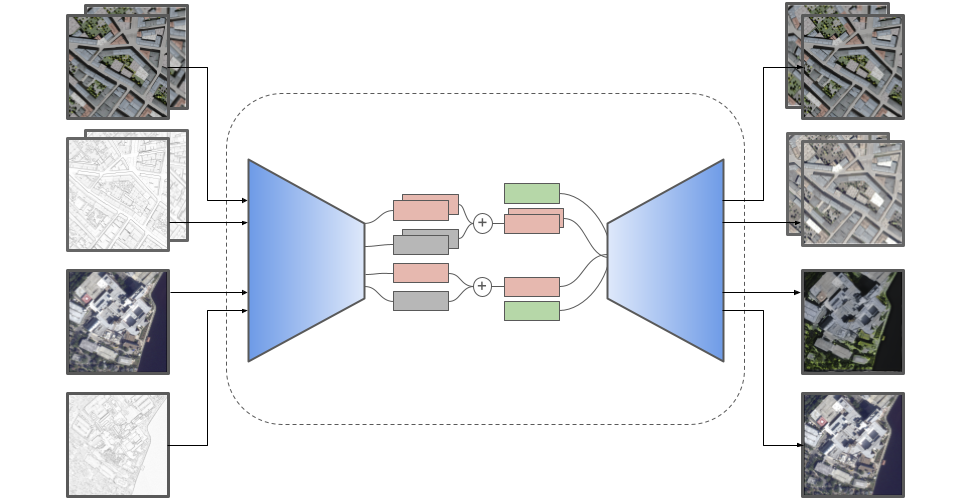}
      \put (-398,228){\scriptsize{$x_a$}}
      \put (-398,147){\scriptsize{$xe_a$}}
      \put (-398,111){\scriptsize{$x_b$}}
      \put (-398,24){\scriptsize{$xe_b$}}
      \put (-270,186){Generator}
      \put (-355,124){Encoder}
      \put (-170,124){Decoder}
      \put (-255,152){\scriptsize{$c_a$}}
      \put (-255,133){\scriptsize{$e_a$}}
      \put (-257,120){\scriptsize{$c_b$}}
      \put (-257,100){\scriptsize{$e_b$}}
      \put (-207,163){\scriptsize{$s_a$}}
      \put (-207,133){\scriptsize{$ce_a$}}
      \put (-207,122){\scriptsize{$ce_b$}}
      \put (-207,90){\scriptsize{$s_b$}}
      \put (-104,228){\scriptsize{$\hat{x}_{aa}$}}
      \put (-104,148){\scriptsize{$\hat{x}_{ab}$}}
      \put (-104,113){\scriptsize{$\hat{x}_{ba}$}}
      \put (-100,23){\scriptsize{$\hat{x}_{bb}$}}
    \caption{Illustration of the generator architecture in an autoencoder with edge map integration. The image along with its corresponding edge map is encoded and added together as content edge code before applying it as an input to the decoder. The decoder merges the content-edge code with style code from every domain to generate content that is contextually fitting. $xc_a$, $xc_b$ represents the input images from both domains (content), $xe_a$, $xe_b$ represents the corresponding edge maps. $c_a$, $c_b$, $e_a$, $e_b$ represents the content and edge code from encoder for both domains. $s_a$, $s_b$ are the randomly initialized style code before the training. $x_{aa}$, $x_{ab}$, $x_{ba}$, $x_{bb}$ represents the respective output images from the decoder. }
    \label{fig:generator_arch}
 \end{figure*}

We address this problem using a  \textit{lightweight edge-aware GAN network}, that performs unpaired image-to-image translation while maintaining the disparity values. At first, the edge maps of input images are obtained from Sobel operator and are provided as an additional input along with image pairs from both domains to the generator. The encoder of the generator computes the content and edge code separately from the input image and its edge map and is added together as content edge code. The content edge code is provided to the decoder along with a random style code to generate images of different domain as shown in \cref{fig:generator_arch}. The use of edge maps ensures that the structure of the image is retained and not lost in translation and thus prevents the matching algorithm fail due to blurred boundaries. Additionally, we use a warping loss, where we warp the left translated image with its respective disparity map and compare it to the right translated image to enforce stereo consistency. Extensive experiments across multiple datasets demonstrate our method outperforms the existing methods quantitatively and qualitatively. Moreover, we use a single lightweight network to perform optimization on two tasks without the use of any pre-trained networks. 
To sum up, our main contributions are:
	
\begin{itemize}
    \item Developing a lightweight framework for \textit{image-to-image translation of stereo pairs} considering a consistent translation of left and right images that preserves the matching. By including edge maps and a warping loss, we improve the matching features of the generated pairs.
    \item Results show that the quality of the translated images leads to a better disparity prediction than other state-of-the-art translation methods. 
    \item By testing on data of remote sensing and autonomous driving tasks, we demonstrate that our approach works with a variety of datasets.
\end{itemize}

\section{Related work}

 \begin{figure*}[ht!]
    \begin{subfigure}{0.20\textwidth}
        \includegraphics[width=\textwidth]{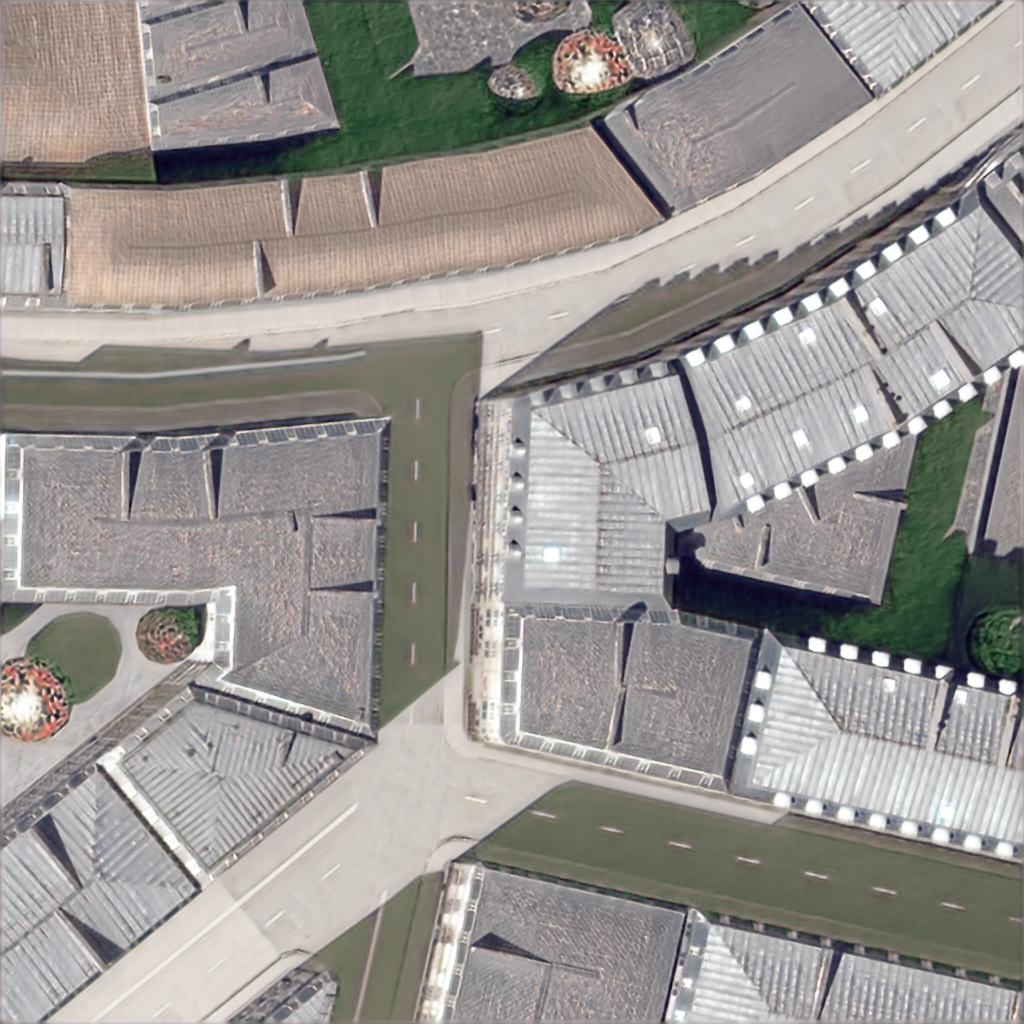}
        \caption{}\label{fig:left1}
    \end{subfigure}
    \hfill %
    \begin{subfigure}{0.20\textwidth}
        \includegraphics[width=\textwidth]{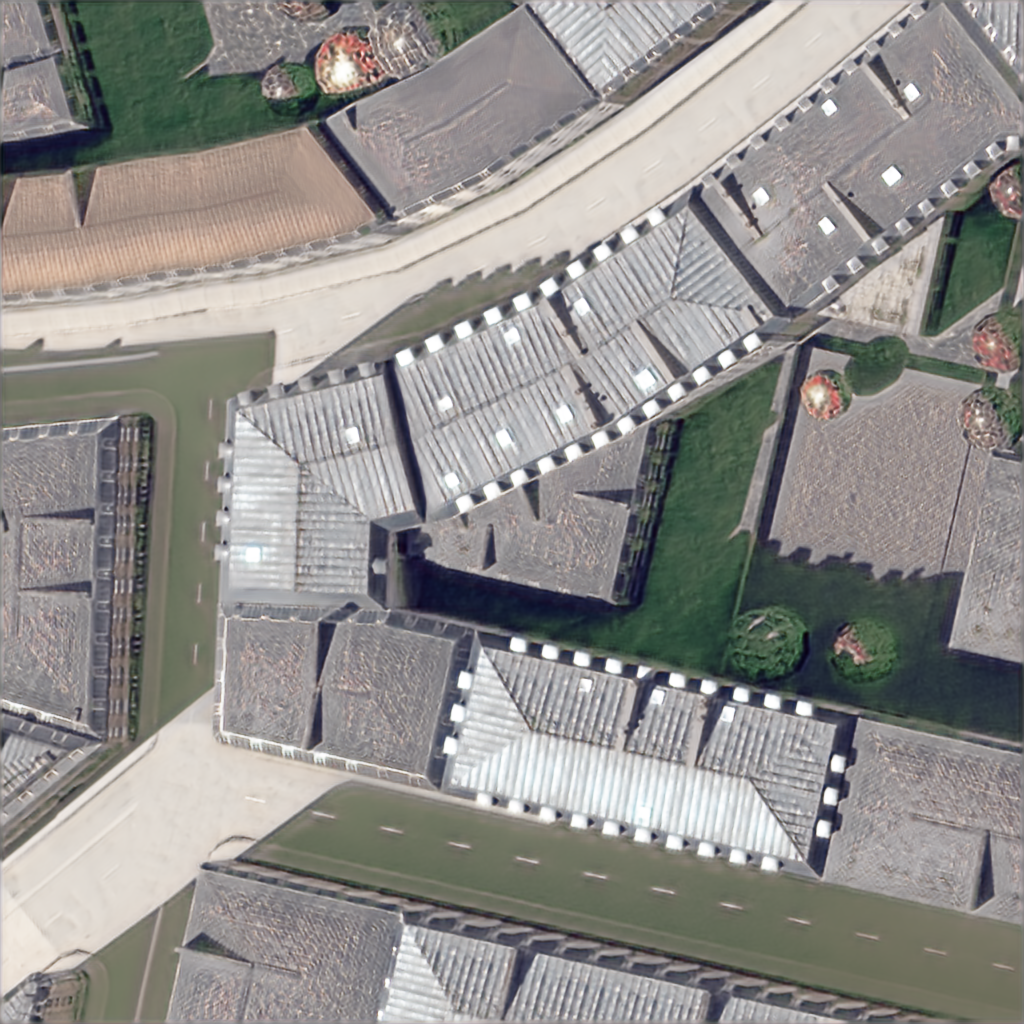}
        \caption{}\label{fig:right1}
    \end{subfigure}
    \hfill %
    \begin{subfigure}{0.20\textwidth}
        \includegraphics[width=\textwidth]{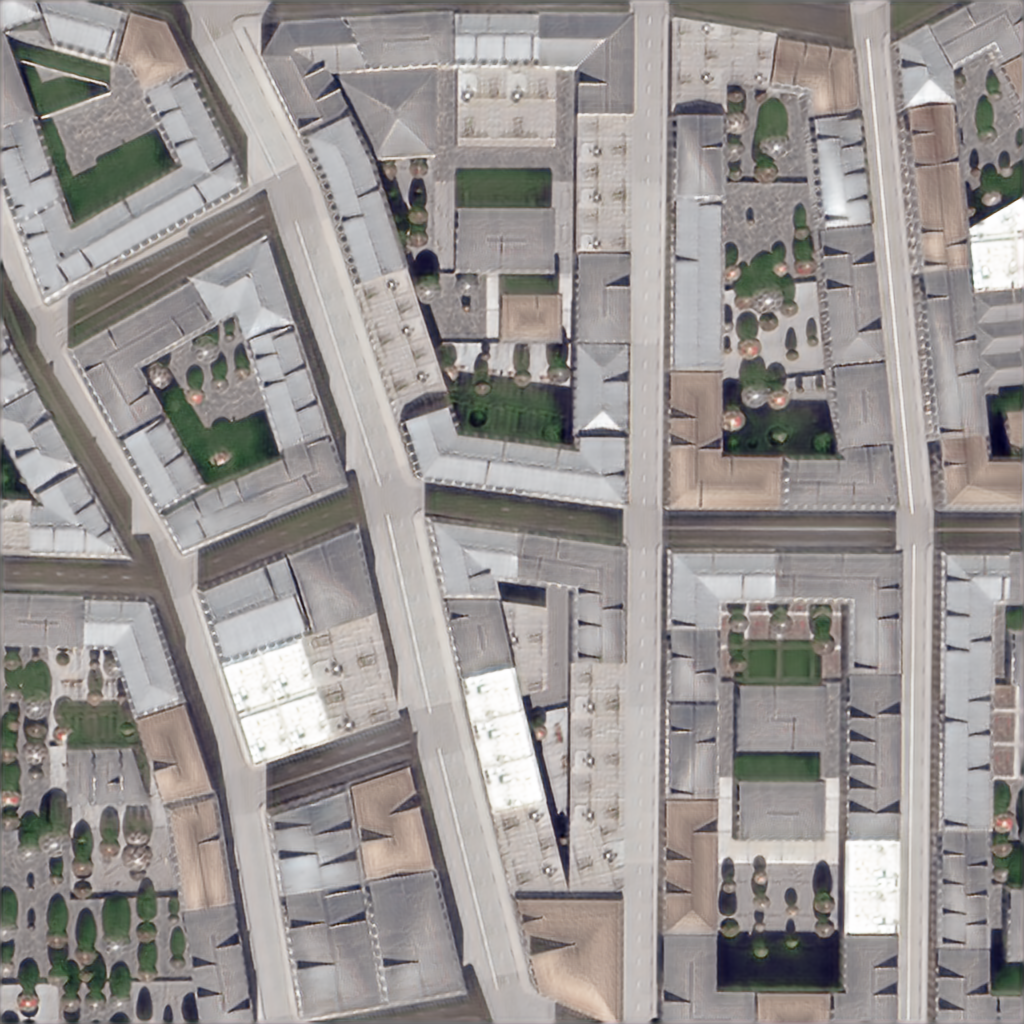}
        \caption{}\label{fig:left2}
    \end{subfigure}
    \hfill %
    \begin{subfigure}{0.20\textwidth}
        \includegraphics[width=\textwidth]{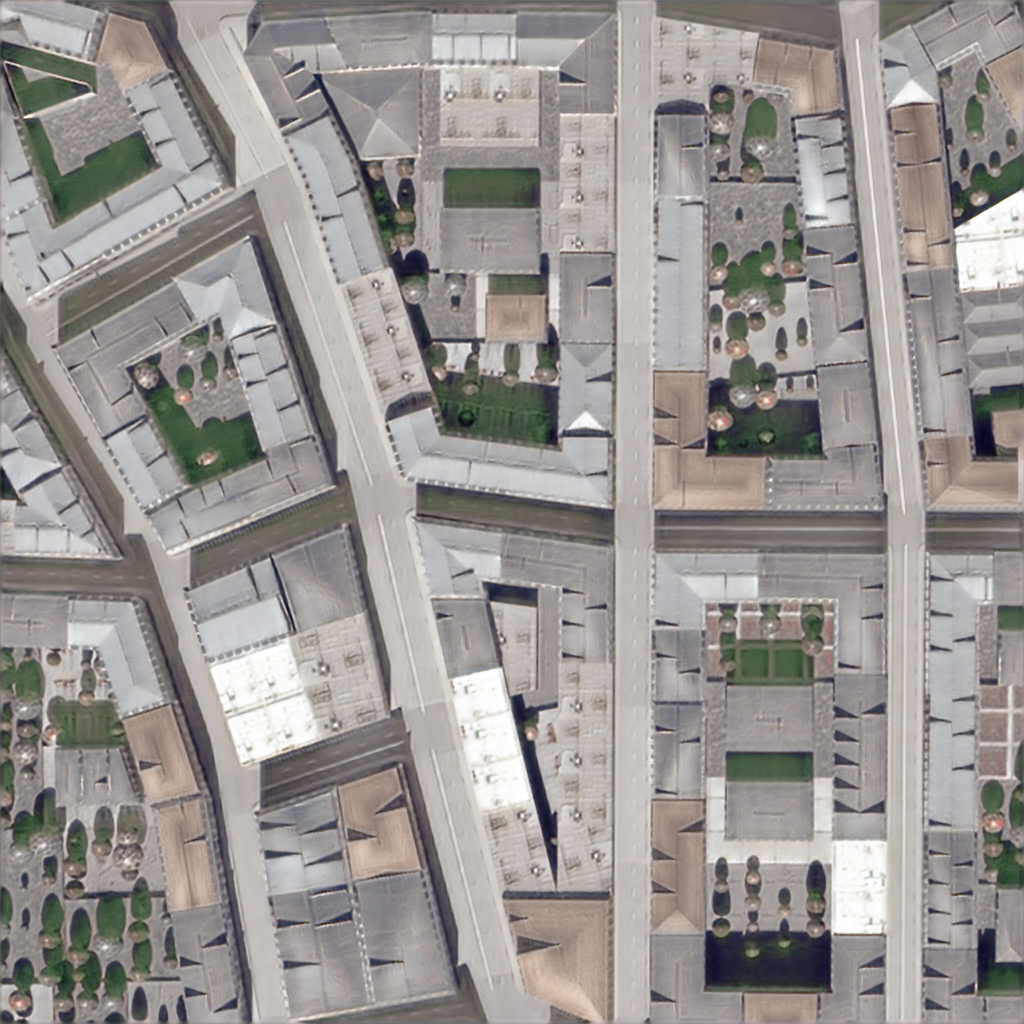}
        \caption{}\label{fig:right2}
    \end{subfigure}
    \hfill %
      \caption{Pairs of translated images. For the translated left-view images \ref{fig:left1} and \ref{fig:left2}, the corresponding right-view images \ref{fig:right1} and \ref{fig:right2} are also displayed. As can be seen from the images, a semantic-consistent translation is applied to both the left and right-view images.}
\end{figure*}
\subsection{Stereo matching}

Semi-global matching~\cite{sgm}, a classical stereo method uses pixel-wise matching cost for computing the disparities between two images. It produces an approximate global optimal solution and is still one of the best performing techniques for disparity estimation in certain domains. MC-CNN~\cite{mccnn} introduced deep learning based techniques to estimate disparity using a deep siamese network where matching cost for cost volume is computed from the network and cost aggregation is carried out through average pooling, followed by additional refinement. GANet~\cite{ganet} has optimized the cost aggregation by introducing semi-global aggregation (SGA - guided) layer which aggregates matching cost in multiple directions and local guided aggregation (LGA) to recover disparities at thin structures and object edges. The use of warping operation as a constraint for stereo matching has proven to be useful. It ensures that the estimated disparities produce geometrically consistent results. Nonetheless, all of these methods suffer from a significant domain gap when applied to data from a different acquisition nature.

Other disparity algorithms focus on tackling the domain gap while learning the matching. CFNet~\cite{shen2021cfnet} proposes a fused cost volume representation in a cascade design for a more robust learning that can be applied to different domains. DSMNet~\cite{zhang2020dsmnet} applies a domain normalization to the input images leading to sharper disparity maps. GrafNet~\cite{liu2022graftnet} transforms the features with a U-shape network before inputting them to the cost volume, improving the domain adaptation. AdaStereo~\cite{song2022adastereo} transforms the color distribution into the target domain while training. In other cases, networks are designed to have a robust domain generalization on unseen data such as RAFT-Stereo~\cite{raftstereo}, which is based on convolutional gated recurrent units (GRUs) or the case of IGEV-Stereo~\cite{igev_stereo}, where geometry encoding improves the results in both stereo and multi-view networks. FC-stereo~\cite{zhang2022fcnet} computes two losses (selective whitening and contrastive feature) to preserve the stereo consistency between images and  helped existing networks to generalize better while training only on synthetic data. Still, we consider that an offline adaptation of the dataset might lead to a good domain generalization without compromising the capabilities of the matching algorithms themselves. 

\subsection{Unpaired image-to-image translation}
Unpaired image-to-image translation translation aims to learn the mapping from a source image domain to a target image domain without paired training data. CycleGAN~\cite{cyclegan} has been a pioneer in solving this task by identifying the key mappings in unpaired data from two different domains. 
The authors introduced cycle consistency loss to constrain the one-to-one mapping space by reconstructing the original image back from the translated image. This loss, in conjunction with adversarial loss and identity loss, plays a pivotal role in image-to-image translation, leading to remarkable visual results. The CUT~\cite{CUT} model extends this concept for one-sided image translation with a contrastive loss. It is calculated using negative samples obtained from the same input, thus enabling faster training. UNIT~\cite{unit} carries out unsupervised image-to-image translation under the assumption that images from both domains consist of a shared latent space. The model uses weight sharing between the layers of generators and discriminators to learn the joint distribution of data. MUNIT~\cite{munit} extends this architecture to handle multiple styles using the disentanglement principle to obtain content and style code separately. The content code from the image is combined with a random style code from cross-domains to obtain diverse styled images. However, these GAN-based methods produce visually appealing image translations but sometimes fail to maintain semantic consistency between source and translated image especially given large number of features. In certain domains such as remote sensing as shown in \cref{fig:Figure2}, CUT~\cite{CUT} still suffers from the problem of hallucination.

\subsection{Synthetic-to-real translation}
The task of translating synthetic images to realistic while preserving semantic consistency has been an active research topic with multiple applications such as semantic segmentation, stereo matching and pseudo label learning. A straightforward approach relies on auxiliary information which is extracted from a task network to track changes in the source and target domains. CyCADA~\cite{hoffman2018cycada} leverages a method that preserves semantic consistency
by constraining on a cycle consistent task-loss. It uses an additional loss which tracks the discrepancy between segmentation maps predicted by a pre-trained segmentation network from the generated images and the ground truth maps. Chen et al.~\cite{chen2019learning} extend the method further by incorporating depth maps. 
Semantic-aware Grad-GAN~\cite{semtic-aware-grad-gan} introduces a soft gradient-sensitive objective and a semantic aware discriminator for domain adaptation of virtual to real urban scenarios.  To address alterations affecting object boundaries in generated images, their method involves applying the Sobel filter to both the image and its corresponding semantic map for deviation tracking. StereoGAN~\cite{stereogan} is specifically designed for the task of translating synthetic images to realistic domain while maintaining the stereo constraints. It utilizes a CycleGAN for image translation and a DispNet~\cite{dispnet} for disparity estimation. SDA~\cite{sda} utilizes the spatial feature transform to fuse features of edge maps with source images. Different than the previous works, Secogan~\cite{secogan} utilized content disentanglement architecture from MUNIT for translating synthetic images of autonomous driving datasets to realistic domain. Instead of relying on a task network, it performs content disentanglement by employing fixed style codes in the generator, making the model computationally effective while preserving semantic consistency.

The task of translating images to a realistic style while maintaining the content structure for stereo matching is a dual optimization task. Although the existing networks address this issue, they suffer when applied to remote sensing images due to large baselines (which implies more occlusion), different acquisition times for left and right images, and city growing. The models developed are predominantly applied in the field of autonomous driving and struggle with achieving domain generalisation. Another challenge is the training of existing models tends to become computationally expensive, as it is a combination of two deep learning networks, one for image translation and the latter for stereo matching. The number of parameters required for training is high and can slow the training process. We address both of the above concerns in our work by employing a single edge-aware image translation GAN model trained additionally with warping loss to enforce the stereo constraints.

\section{Method}
\begin{figure}
    \includegraphics[width=\linewidth]{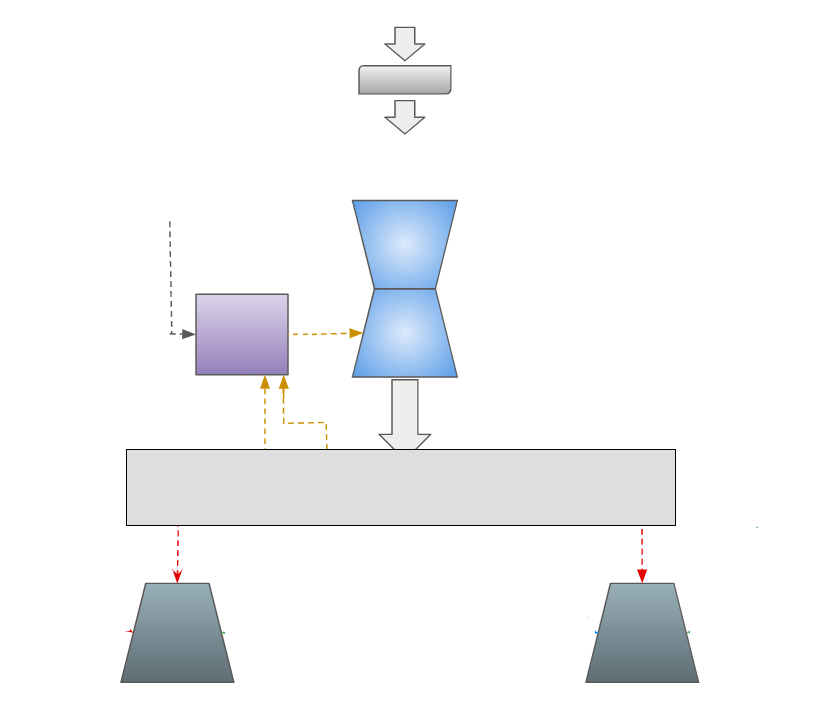}
    \put (-192,20) {\scriptsize{$D_a$}}
    \put (-62,20) {\scriptsize{$D_b$}}
    \put (-176,106) {\scriptsize{Warp}}
    \put (-156,111) {\scriptsize{$\mathcal{L}_{\tiny{warp}}$}}
    \put (-125,130) {\scriptsize{E}}
    \put (-125,106) {\scriptsize{G}}
    \put (-190,142) {\scriptsize{$x_d$}}
    \put (-150,62) {\scriptsize{Generator Output}}
    \put (-130,177) {\scriptsize{Sobel}}
    \put (-170,200) {\scriptsize{$I = \{ {(x_l,x_r,x_d)}_d, x_b, s_a, s_b \} $}}
    \put (-190,155) {\scriptsize{$\{ {(x_l,x_r,x_d)}_a, {(xe_l,xe_r)}_a, x_b, xe_b, s_a, s_b \} $}}
    \caption{Illustration of the GAN-based model architecture featuring multiple loss functions. The design incorporates a combination of adversarial, reconstruction, cycle and warping losses. Adversarial loss promotes realistic image generation, while reconstruction loss ensures faithful reproduction of input data, cycle loss enforces the correct mapping between domains and warping loss enforces geometrical stereo constraints.  }
    \label{fig:arch_2}
\end{figure}

 \begin{figure*}[htp]
\centering
\newcommand{\rulesep}{\unskip\hfill\hfill\hfill\hfill\hfill\hfill\hfill\hfill\hfill\hfill\hfill\hfill\hfill\hfill\hfill\hfill\hfill\hfill\hfill\hfill\hfill\hfill\hfill{\color{white}\vrule}\hfill\ignorespaces}
\begin{subfigure}{.48\linewidth}
\centering
\begin{tabular}{p{0.01cm}ccc}
\rotatebox{90}{\phantom{blan}Synthetic} &
\includegraphics[width=.3\linewidth]{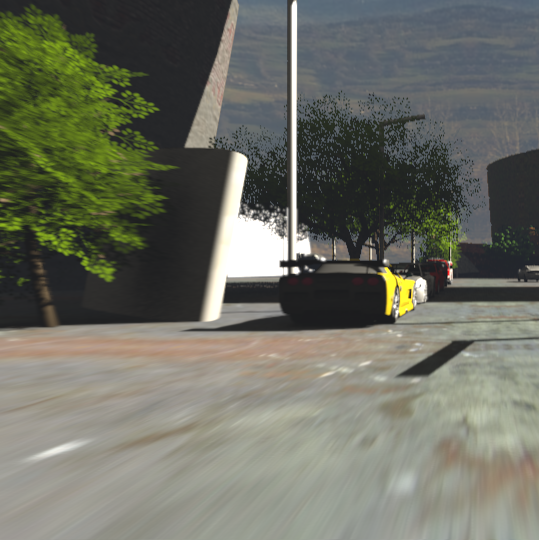} & \includegraphics[width=.3\linewidth]{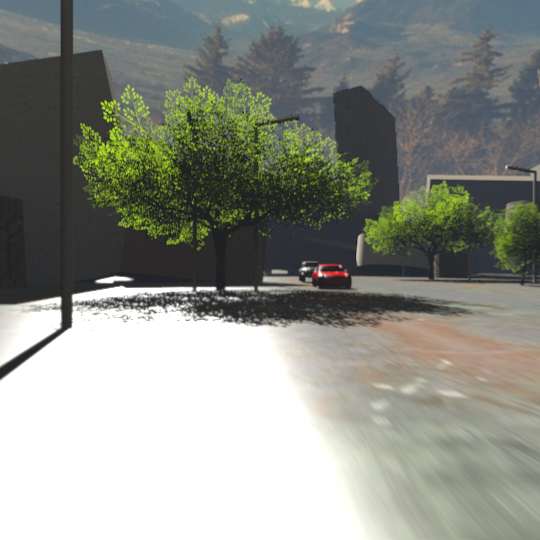} & \includegraphics[width=.3\linewidth]{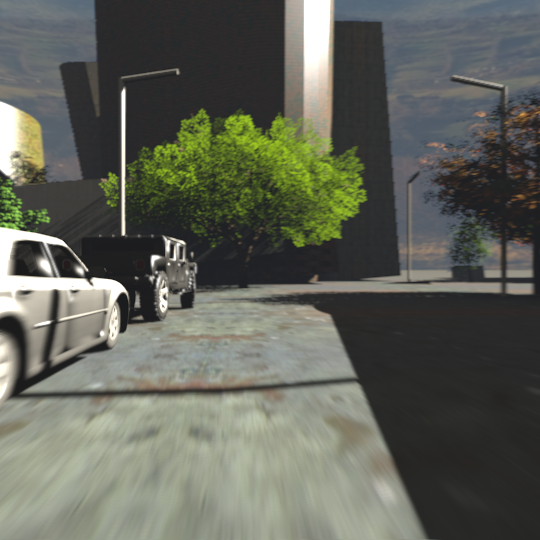} \\
\rotatebox{90}{\phantom{bla}StereoGAN} &
\includegraphics[width=.3\linewidth]{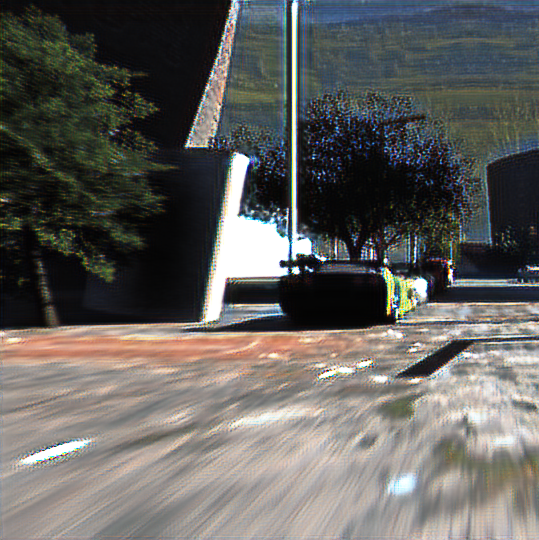} & \includegraphics[width=.3\linewidth]{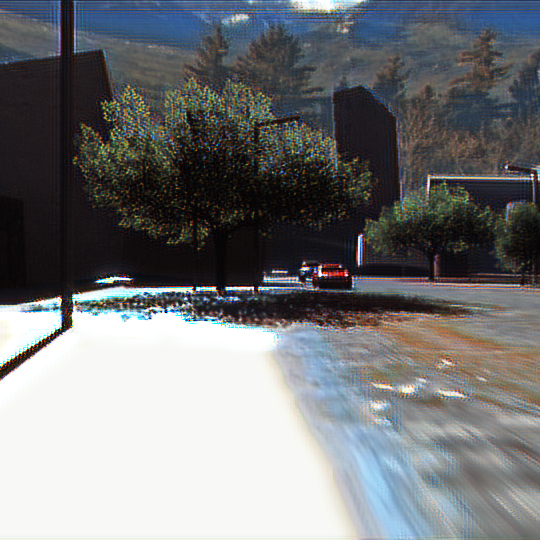} & \includegraphics[width=.3\linewidth]{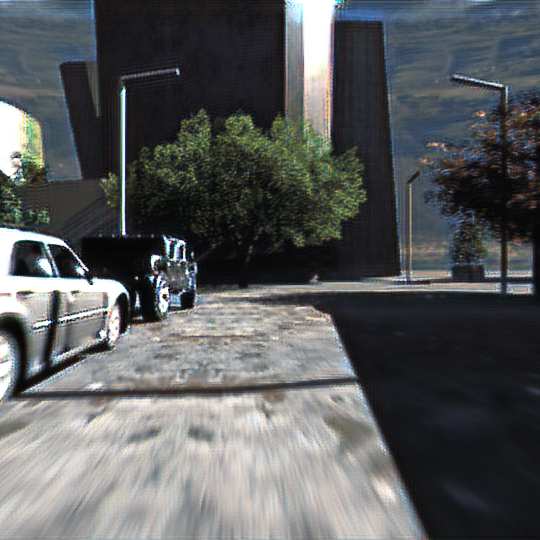} \\
\rotatebox{90}{SyntStereo2Real} &
\includegraphics[width=.3\linewidth]{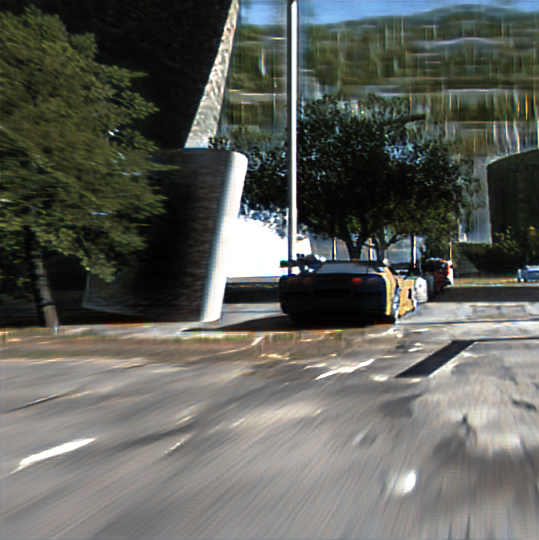} & \includegraphics[width=.3\linewidth]{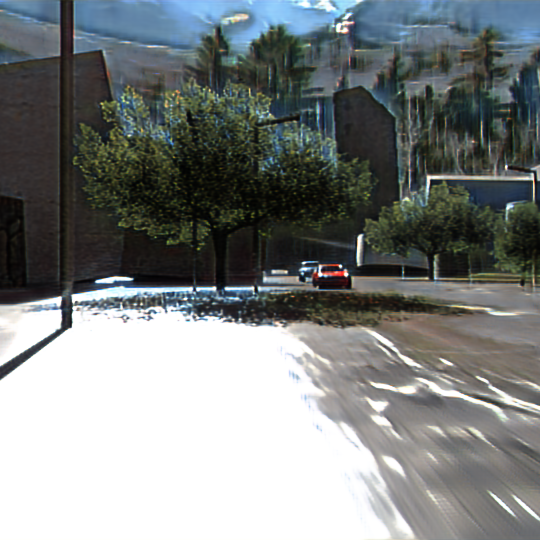} & \includegraphics[width=.3\linewidth]{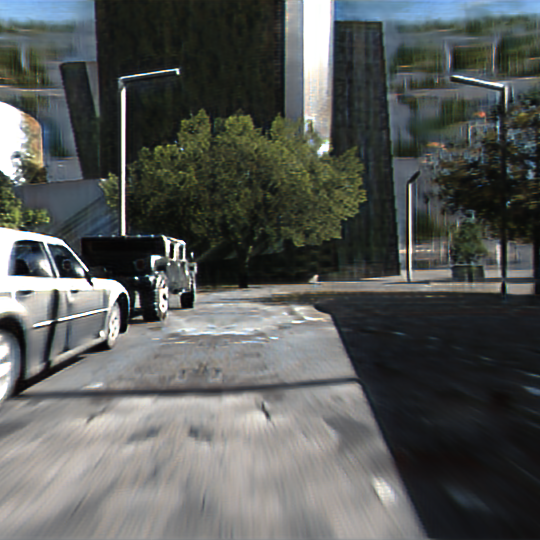}
\end{tabular}
\caption[caption]{Driving to KITTI2015}
\label{fig:fig:driving2kitti}
\end{subfigure}
\rulesep
\begin{subfigure}{.48\linewidth}
\centering
\begin{tabular}{ccc}
\includegraphics[width=.3\linewidth]{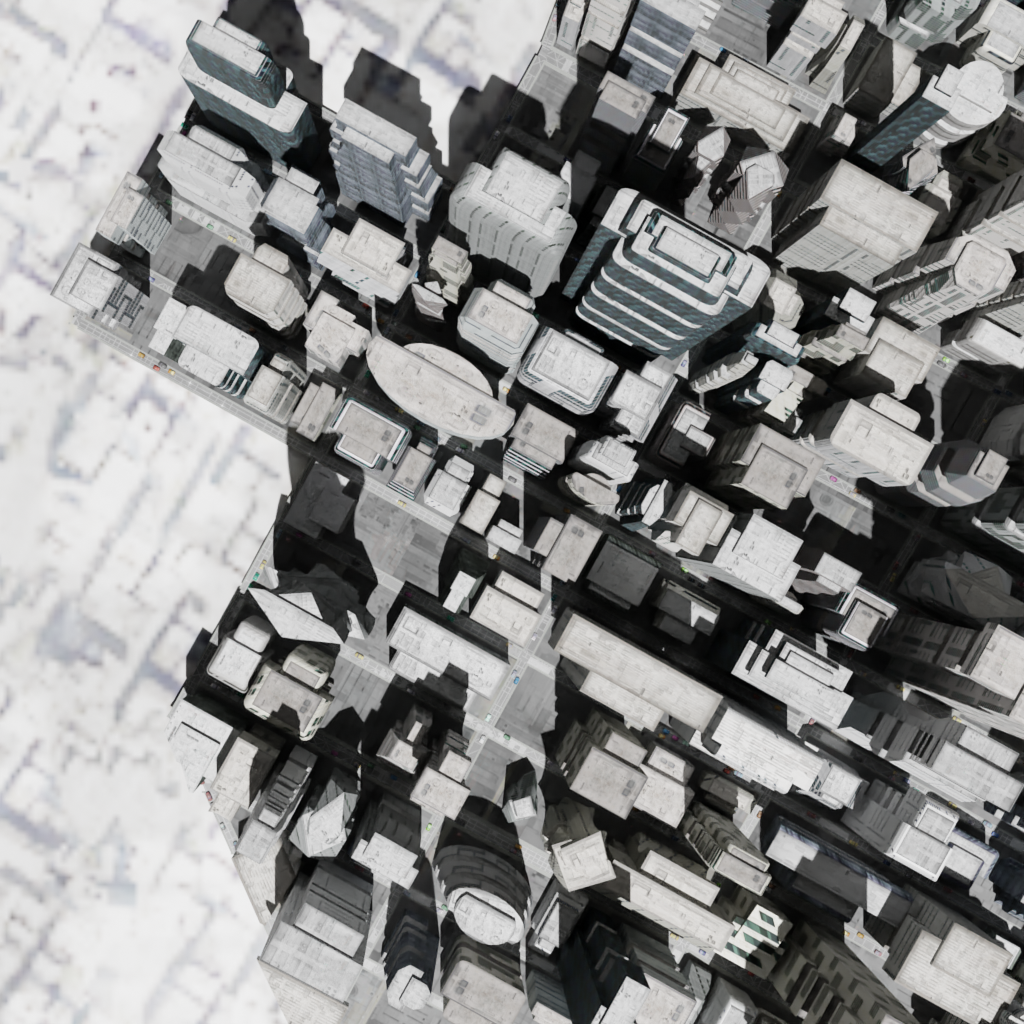} & \includegraphics[width=.3\linewidth]{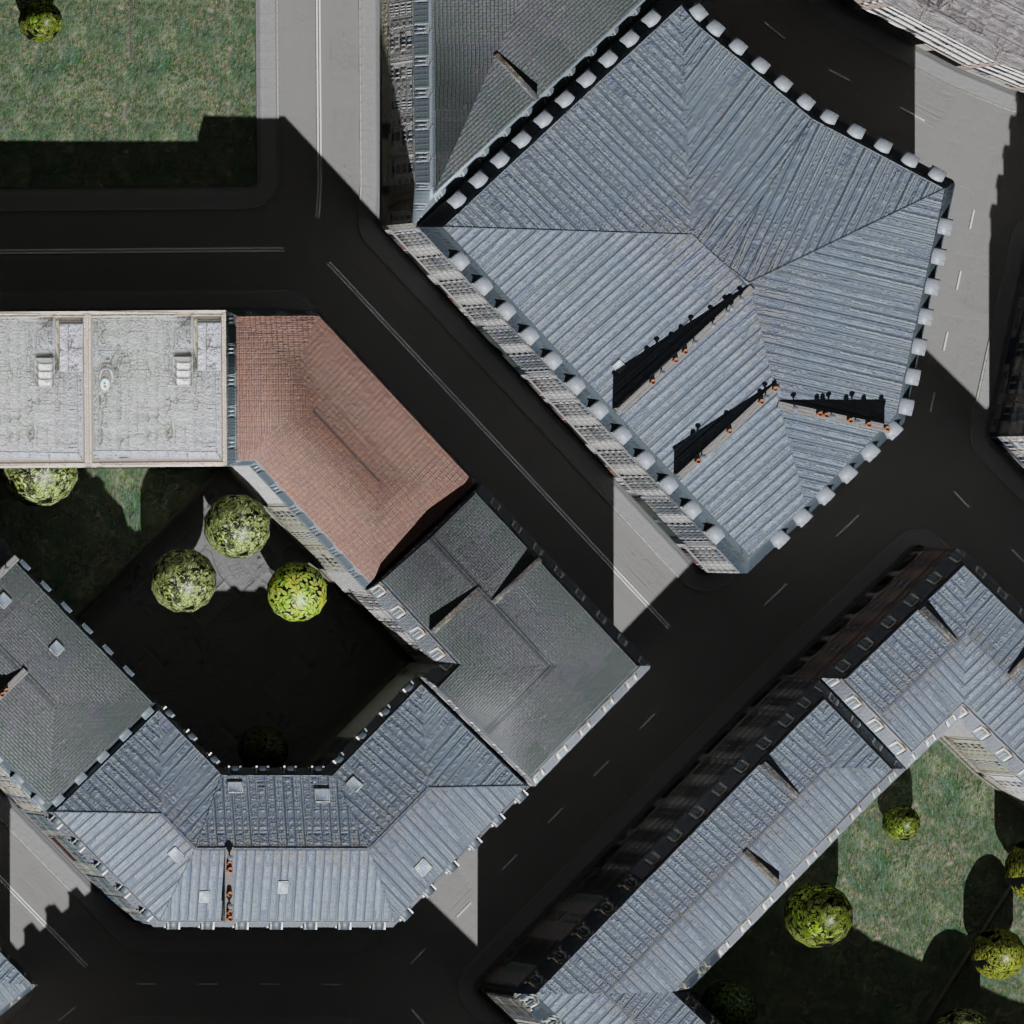} & \includegraphics[width=.3\linewidth]{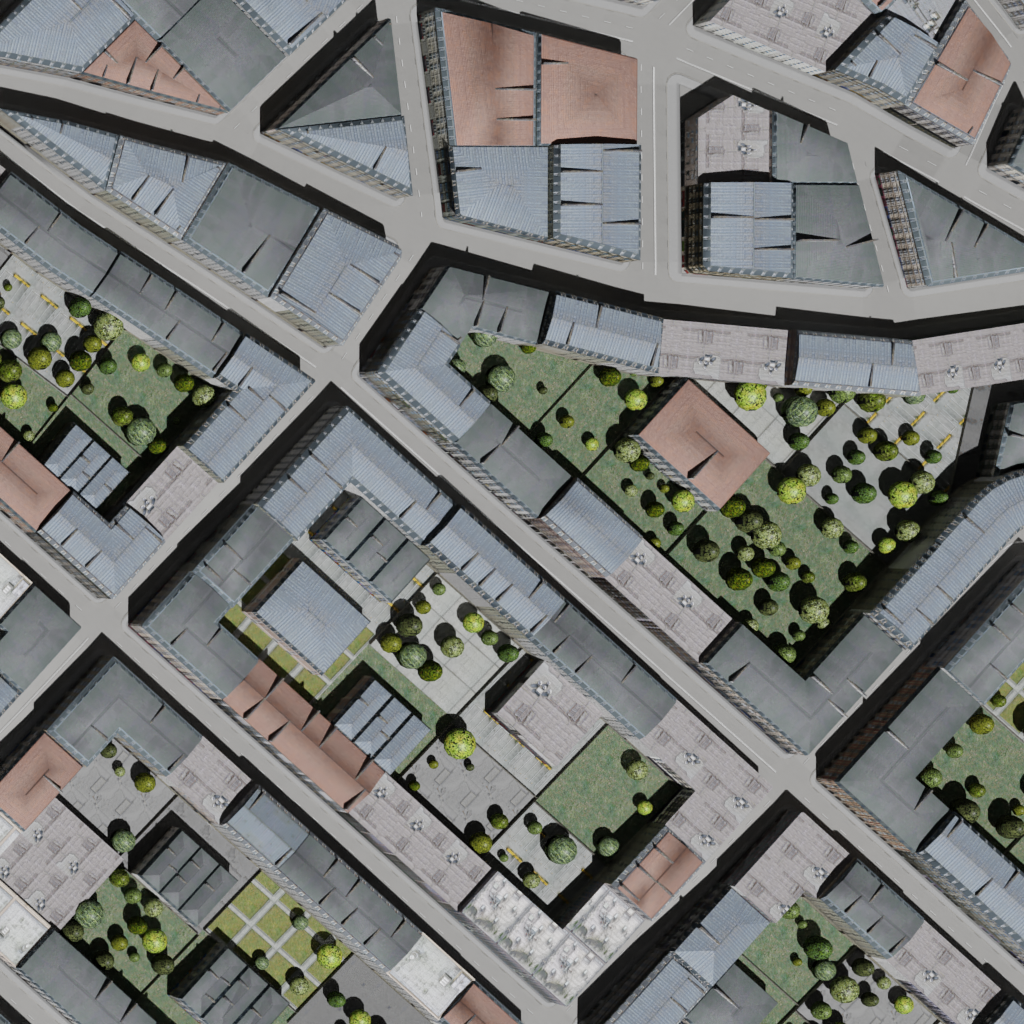} \\
\includegraphics[width=.3\linewidth]{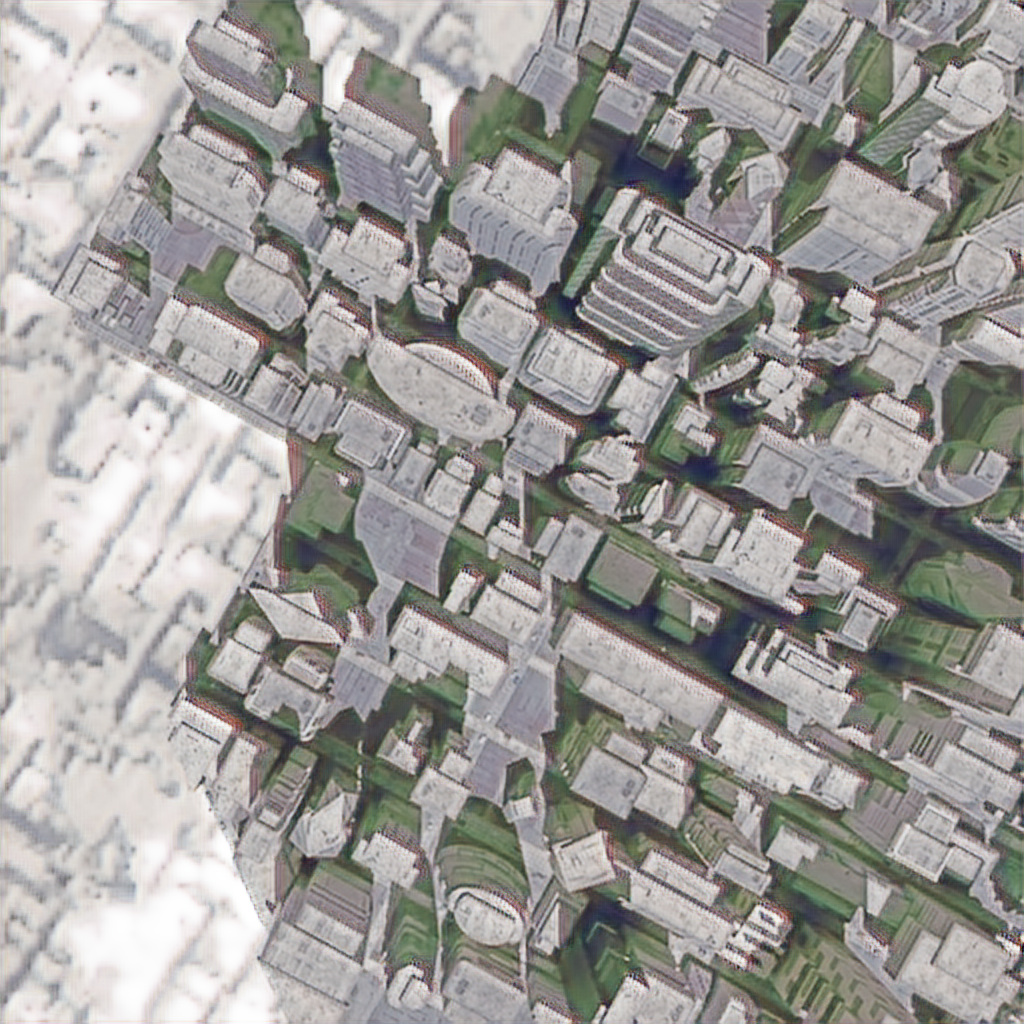} & \includegraphics[width=.3\linewidth]{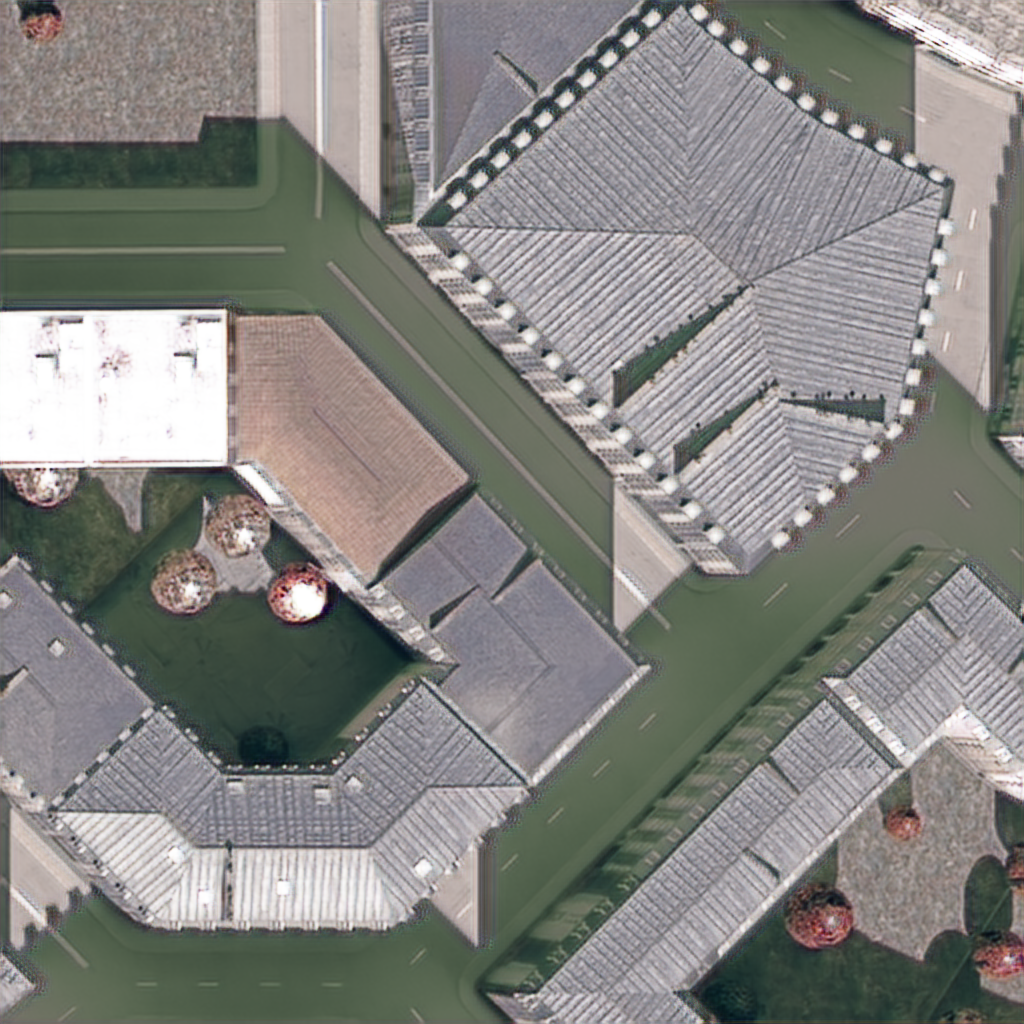} & \includegraphics[width=.3\linewidth]{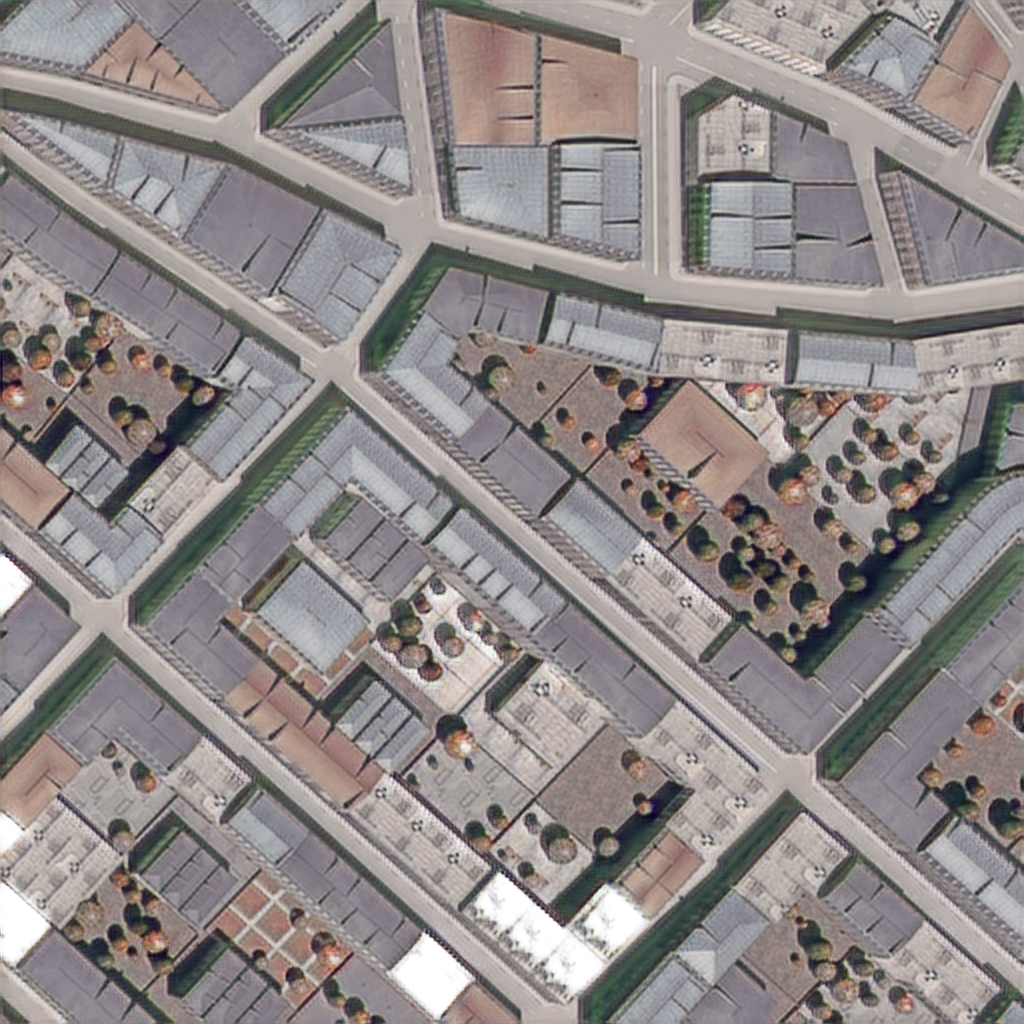} \\
\includegraphics[width=.3\linewidth]{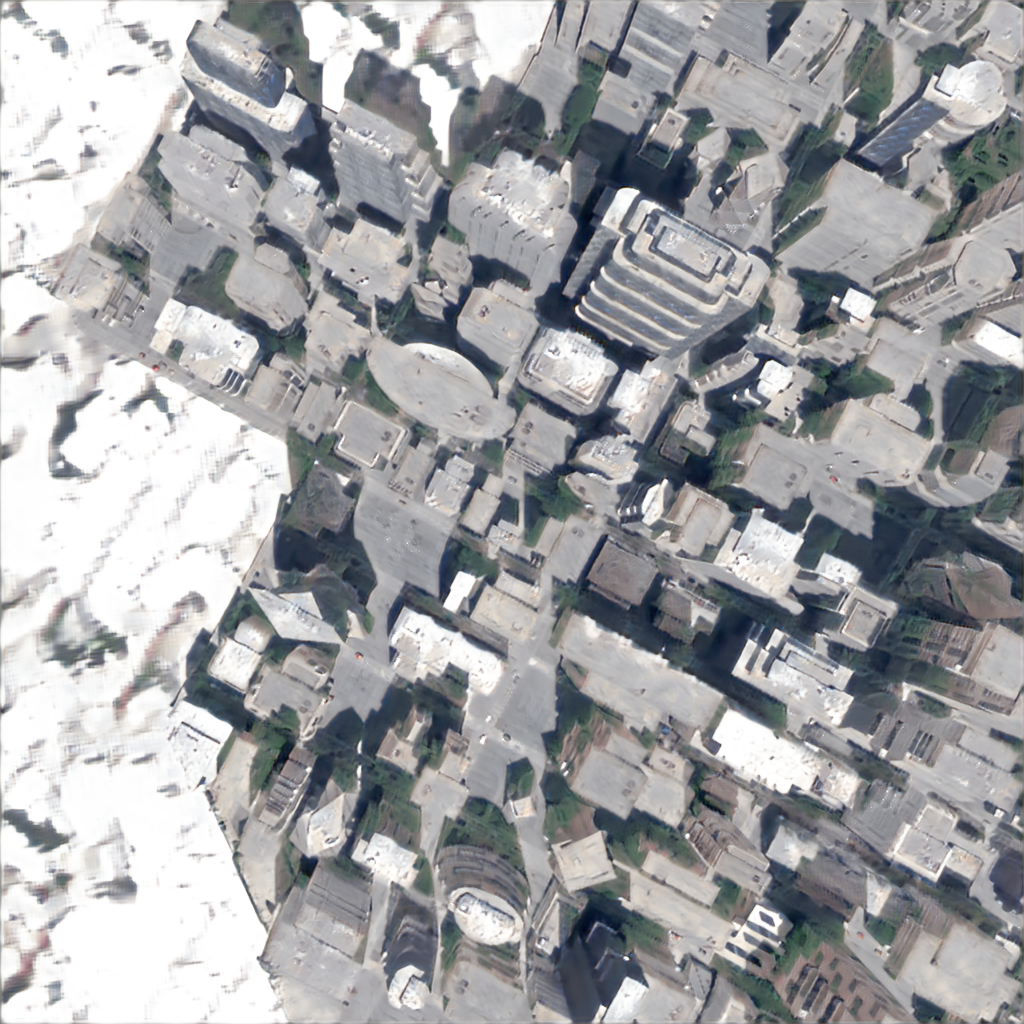} & \includegraphics[width=.3\linewidth]{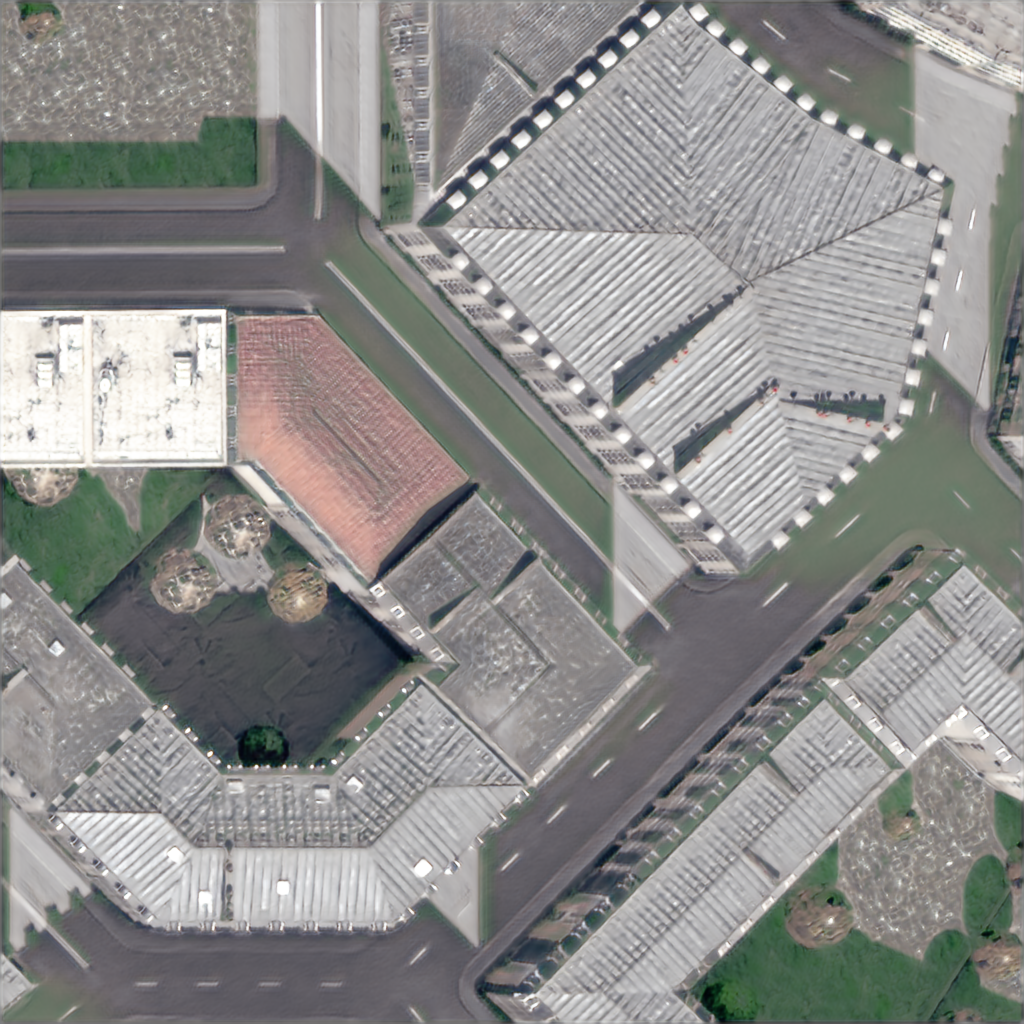} & \includegraphics[width=.3\linewidth]{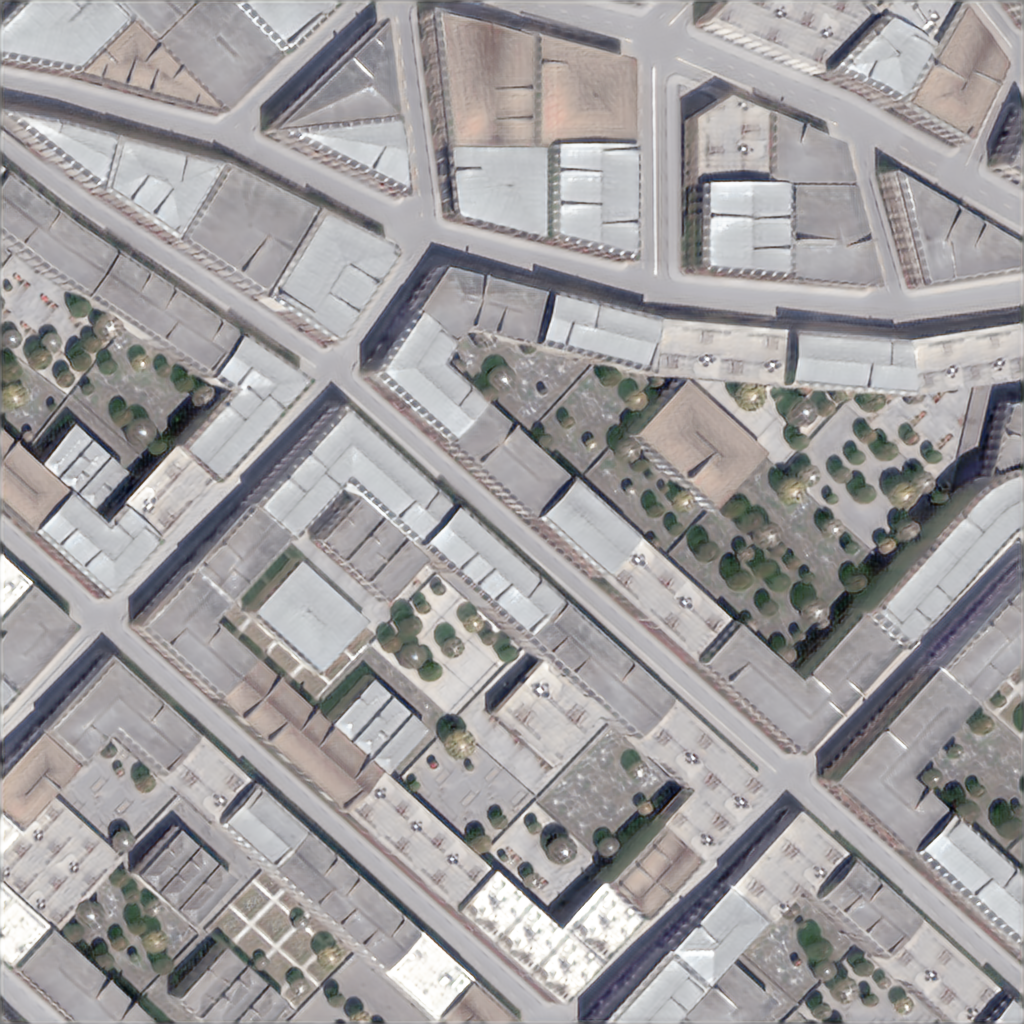} \\
\end{tabular}
\caption[caption]{SyntCities to US3D}
\label{fig:fig:synt2real}
\end{subfigure}

\caption{Comparison of image translations: The first row showcases original synthetic images, the second row presents images translated
using StereoGAN, and the third row exhibits images translated using our SyntStereo2Real.}
\label{fig:results}
\end{figure*}

We carry out the translation of synthetic to realistic domain images under the assumption that both domains share universal features that describe the elements in the scene (such as buildings, roads, vegetation), as well as distinctive features specific to the particular domain, focusing on visual attributes like appearance or style.

Given a synthetic \textit{left-right-disparity} tuple ${(x_l,x_r,x_d)_a}\in\mathcal{X}_a$ denoting the stereo pair of left and right image with its corresponding disparity for source domain, a real image ${x_b} \in \mathcal{X}_b$ representing the target domain, and two randomly sampled style codes $s_a$, $s_b$ for each domain, our model synthesizes a realistic stereo matched pair of the synthetic image. 

Our work draws inspiration from MUNIT~\cite{munit} and Secogan~\cite{secogan} to learn disentangled representations from two domains without supervision.  Similar to ~\cite{munit}, our translation model consists of an autoencoder (encoder $E$ and decoder $G$) as a generator for both domains. The encoder factorizes each input into latent content code $c_i (i=a,b)$, where $c_i = E({x}_i)$. Style code is initialized before the training using normal distribution  as $s_i = (\gamma_i, \beta_i)$~\cite{secogan}, for each domain and remains constant during the training. Here $\gamma_i$ and $\beta_i$ represents the mean and standard deviation of the normal distribution. Edge maps of the corresponding input images are obtained from the Sobel operator ${xe}_i = SO({x}_i)$ and are given as additional input to preserve structural information. The encoder generates the latent edge code $e_i = E({xe}_i)$ from the edge maps. The edge code is added to the content code as content-edge code ${ce}_i = c_i + e_i$ and is provided as an input to the decoder as shown in \cref{fig:generator_arch}. The decoder generates the output image by swapping the content and style codes. The discriminator distinguishes the original image to the generated image by adversarial training. Since we have a real and synthetic domain, we have two discriminators $D_A$ and $D_B$.

 Multiple losses help in constraining and generating images in a meaningful manner in GAN based networks. \Cref{fig:arch_2} shows an overview of the losses used in the training of the model.  A reconstruction loss:
\begin{equation}
\mathcal{L}_{rec}^{aa} (E, G) =  E_{x_a \sim X_a} \Vert G(E(x_a),s_a) - x_a\Vert_1,
\label{reconstruction_loss}
\end{equation}
ensures that the model generates accurate reconstruction of images after content disentanglement.

In image-to-image translation, it is essential that the generated images in the target domain are not only realistic but also faithfully represent the original content. Cycle consistency loss~\cite{cyclegan}: 
\begin{equation}
\mathcal{L}_{cycle}^{aba} (E, G) =  E_{x_a \sim X_a} \Vert G(E(x_{ab}),s_a) - x_a\Vert_1,
\label{cycle_loss}
\end{equation}
enforces this constraint by calculating the loss between original image and the transformation of original image to another domain ($x_{ab}$), and transform it back again to original domain ($x_{aba}$).

Since we use a GAN based approach to train the model, we use an adversarial loss:
\begin{multline}
    \begin{aligned}
        \mathcal{L}_{adv}^{a} (E, G, D_a) =  &E_{x_a \sim X_a} logD_a(p(x_a)) \: + \\ &E_{x_b \sim X_b} log(1 - D_a(p(x_{ba})),
        \label{adv_loss}
    \end{aligned}    
\end{multline}
which matches the data distribution of translated images to the distribution of target domain. The adversarial loss is employed by both the discriminator and generator, whereas the other mentioned loss exclusively guides the training of the generator. Since we use a patch based discriminator, the $p$ in \cref{adv_loss} refers to random patches of image.
 
Considering the images from one domain are synthetically generated, we assume to have access to additional information like ground truth labels, disparity maps, and segmentation masks. Warping loss as an additional constraint can be a useful addition, especially in tasks where the images are later used for training disparity estimation models. We compute the warping loss as

\begin{equation}
    \begin{aligned}
        \mathcal{L}_{\text{warp}} = \lambda_1 \cdot \mathcal{L}_{1}(&G(E(x_{r_a}),s_b) \\
        &- W(G(E(x_{l_a}),s_b), x_d) \\
        &+ \lambda_2 \cdot (1 - \text{SSIM}(G(E(x_{r_a}),s_b) \\
        &- W(G(E(x_{l_a}),s_b), x_d))),
    \end{aligned}
\label{warp_loss}
\end{equation}
which compares the warped left image $W(x_{l_{ab}}, x_d)$, which has undergone translation, and the right image after translation $x_{r_{ab}}$. We use a combination of $\mathcal{L}_1$ loss and SSIM loss for calculating the warping loss.

The corresponding losses from other domain $\mathcal{L}_{rec}^{bb}$, $\mathcal{L}_{cycle}^{bab}$ and $\mathcal{L}_{adv}^{b}$ are calculated in a similar manner. Therefore, the overall loss function for the generator is given by

\begin{multline}
    \begin{aligned}
       \min_{E,G} \max_{D_a,D_b}\mathcal{L}(E, G, D_a, D_b) =  &\lambda_3 \cdot (\mathcal{L}_{rec}^{aa} + \mathcal{L}_{rec}^{bb}) \:+ \\ &\lambda_4 \cdot (\mathcal{L}_{cyc}^{aba} + \mathcal{L}_{cyc}^{bab}) \:+ \\ &\lambda_5 \cdot (\mathcal{L}_{adv}^{a} + \mathcal{L}_{adv}^{b}) \:+ \\ &\mathcal{L}_{warp} . 
       \label{gen_loss}
    \end{aligned}
\end{multline}

\section{Experiments}

\begin{table*}[t!]
\centering
\begin{tabular}{lcccc}
\toprule
\textbf{Datasets} & \textbf{Metrics} & \textbf{Baseline} & \textbf{StereoGAN} & \textbf{SyntStereo2Real(ours)} \\
\midrule
\textbf{SyntCities} & MAD $\downarrow$ &  1.801 & 1.520 & \textbf{1.319} \\
 \textbf{to US3D}& 3px-acc\% $\uparrow$ &  63.097 & 66.765 & \textbf{69.906} \\
& 1px-acc\% $\uparrow$ & 30.790 & 33.619 & \textbf{35.928} \\

\midrule
\textbf{Driving} & MAD $\downarrow$ & 0.721 & 0.626 & \textbf{0.575} \\
 \textbf{to KITTI}& 3px-acc\% $\uparrow$ & 88.871 & 89.646 & \textbf{91.373} \\
& 1px-acc\% $\uparrow$ &61.832 & 64.271 & \textbf{65.892} \\

\bottomrule
\end{tabular}
\caption{Comparison of metrics for SyntCities to US3D and Driving to KITTI. The table illustrates the performance across datasets, showcasing results for the SyntCities (baseline), StereoGAN, and SyntStereo2Real (ours). Bold values highlight superior performance in MAD reduction and accuracy enhancement.}
\label{tab:comparison}
\end{table*}

\begin{table}[!ht]
    \centering
    \begin{tabular}{lccccc}
        \toprule
        {}   Model & $n_{params}$ \\
        \midrule
        StereoGAN  & 54M  \\
        SyntStereo2Real(ours)  & 11M  \\
        \bottomrule
    \end{tabular}
    \caption{Comparison of the number of learnable parameters to train model between StereoGAN and SyntStereo2real models.}\label{tab:synthetic-evaluation}
    \label{tab:num_parameters}
\end{table}

\begin{figure}[!ht]
  \subfloat[Reference\label{fig:KITTI_rgb}]{\includegraphics[width=0.48\textwidth]{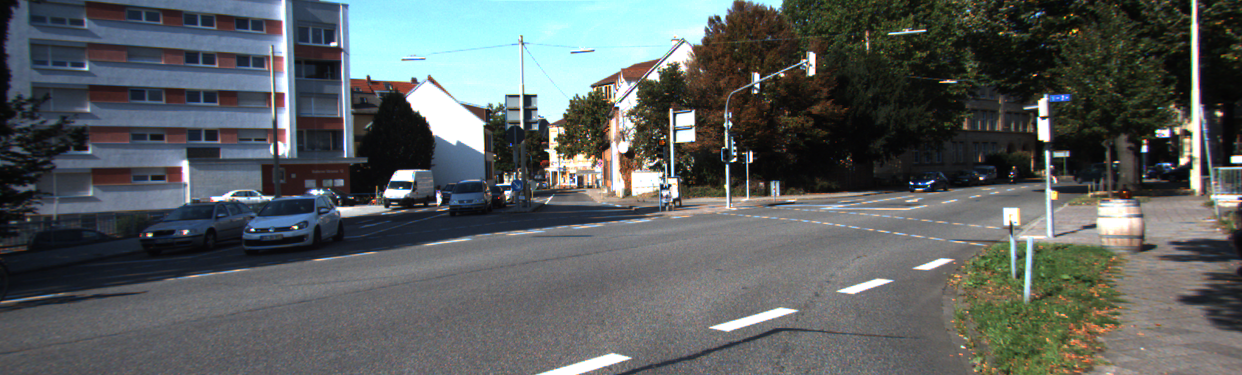}}\\
  \subfloat[Ground Truth\label{fig:fig:KITTI_gt}]{\includegraphics[width=0.48\textwidth]{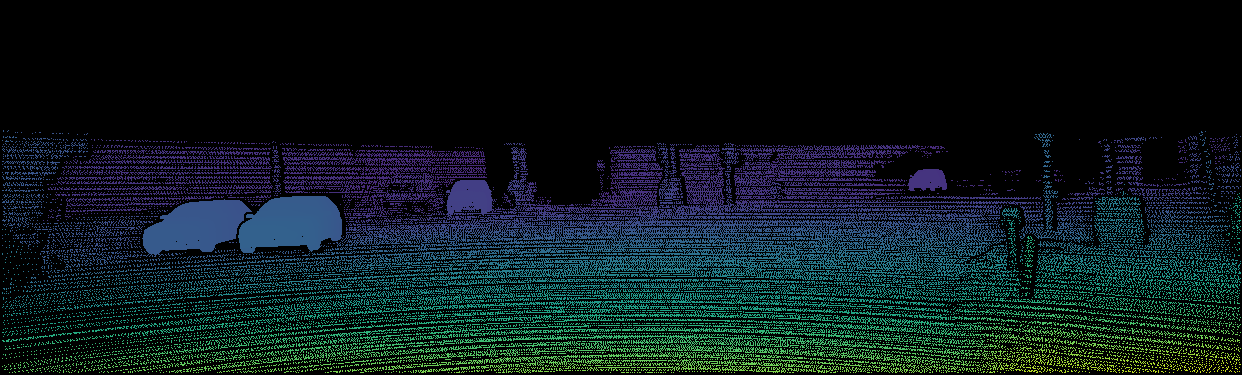}}\\
  \subfloat[Baseline\label{fig:fig:KITTI_ori}]{\includegraphics[width=0.48\textwidth]{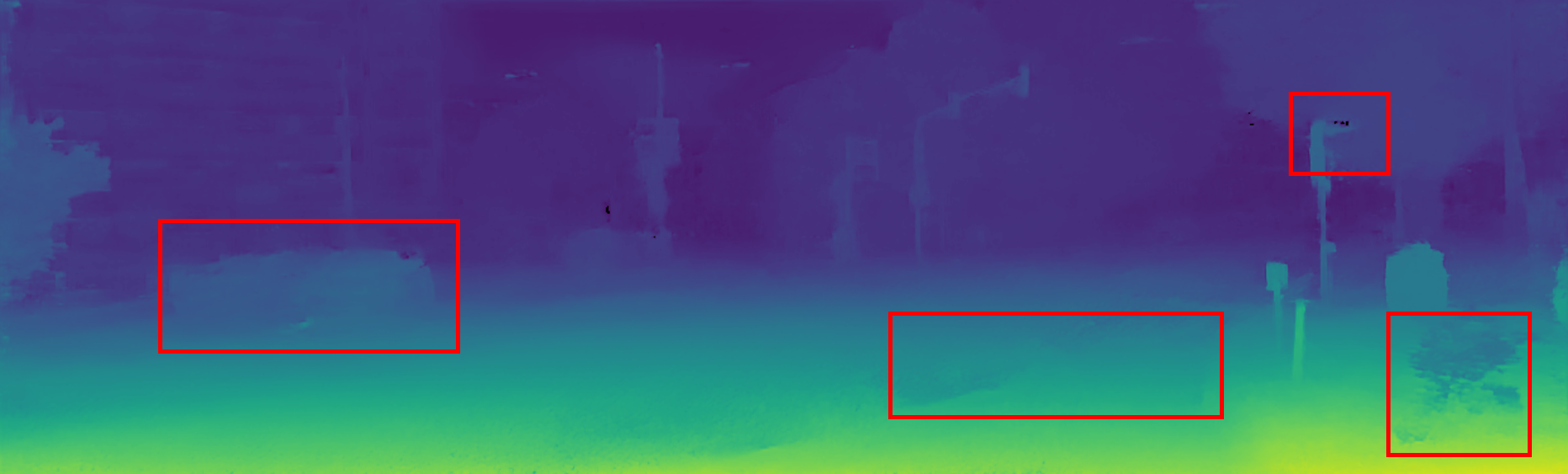}}
  \\
  \subfloat[StereoGAN\label{fig:KITTI_stereogan}]{\includegraphics[width=0.48\textwidth]{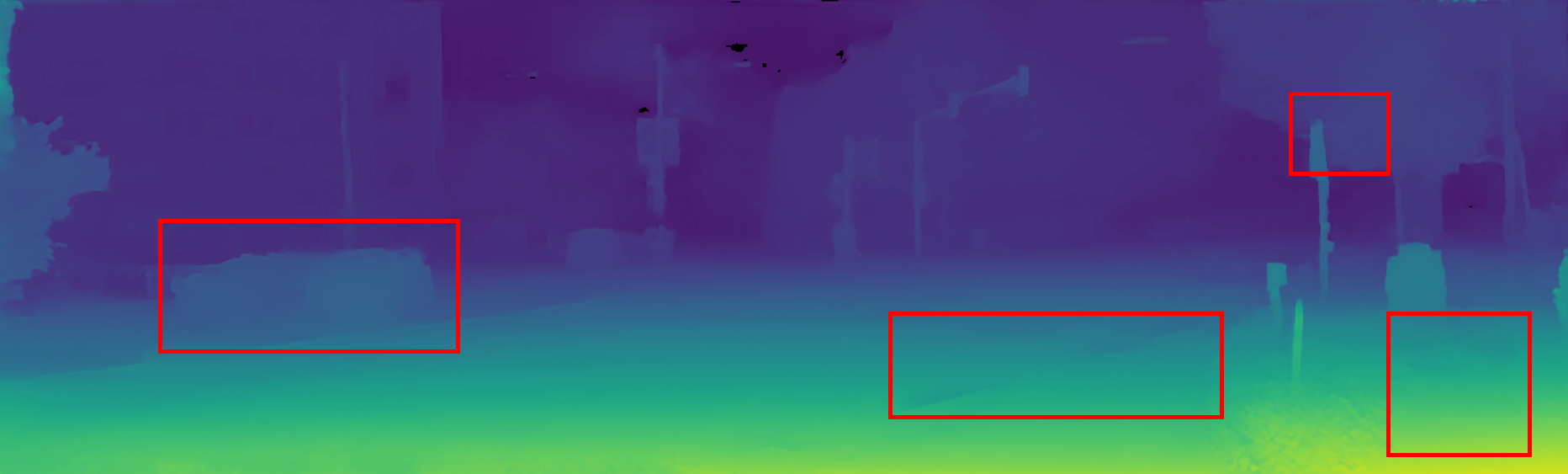}}
  \\
  \subfloat[SyntStereo2Real(ours)\label{fig:KITI_secogan}]{\includegraphics[width=0.48\textwidth]{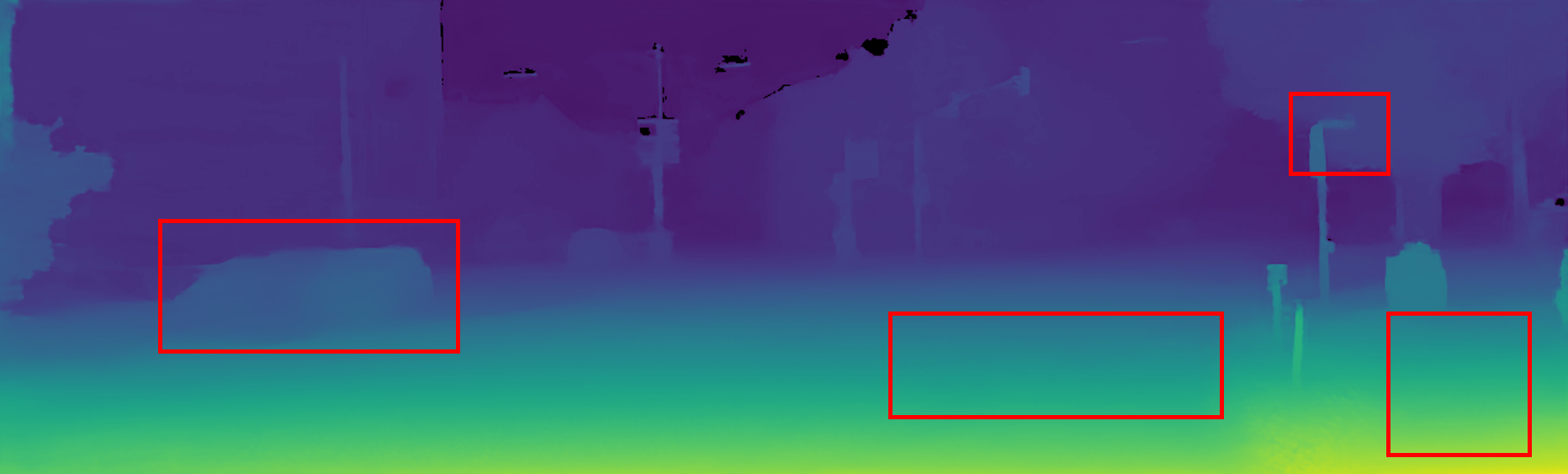}}
  \caption{Results of disparity estimation from the AANet for the KITTI 2015 dataset. Three models are computed for the image shown in (a) RGB reference image, (b) Ground truth, (c) Model trained on Driving (baseline), (d) Model trained on Driving translated using StereoGAN (e) Model trained on Driving translated using SyntStereo2Real (ours).}
  \label{fig:KITTI_disp}
\end{figure}

\begin{figure*}[!h]
  \subfloat[Reference\label{fig:US3D_rgb}]{\includegraphics[width=0.19\textwidth]{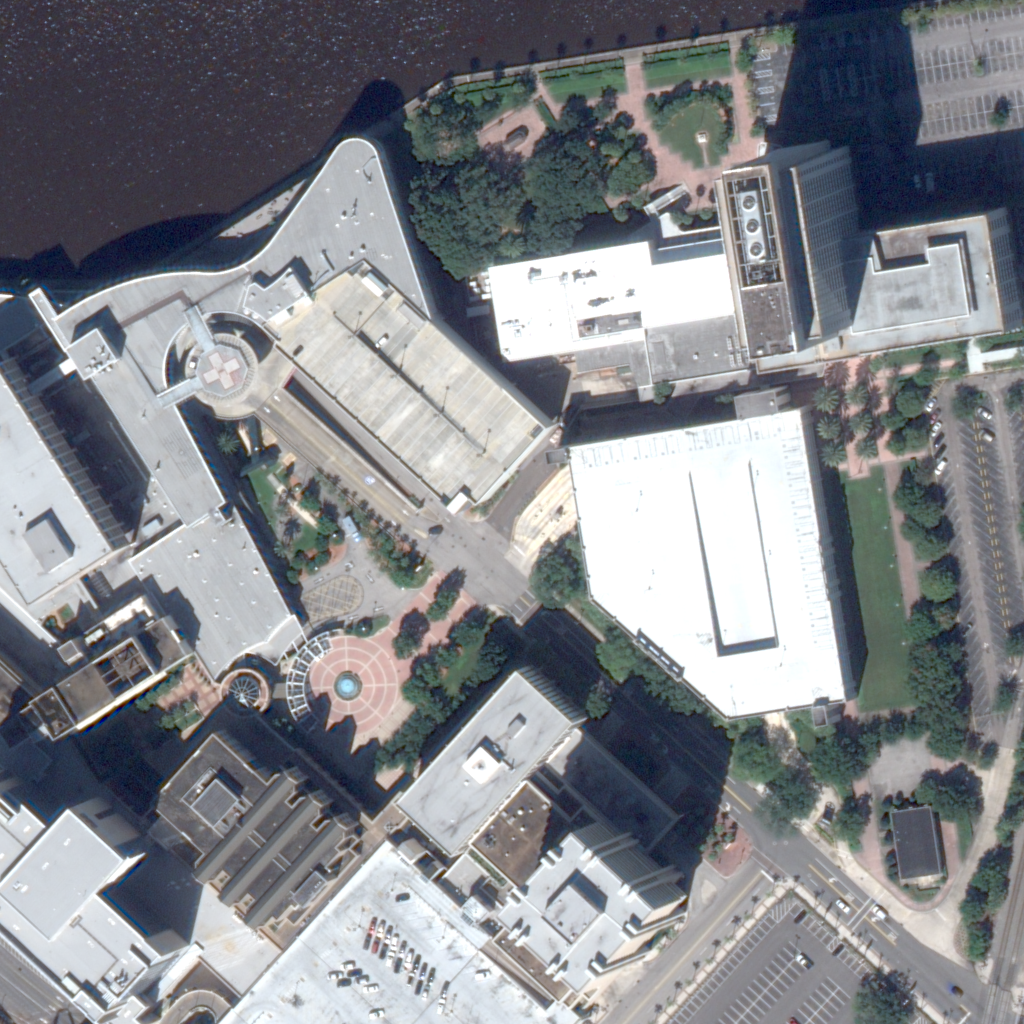}}
  \hfill
  \subfloat[GT\label{fig:fig:US3D_gt}]{\includegraphics[width=0.19\textwidth]{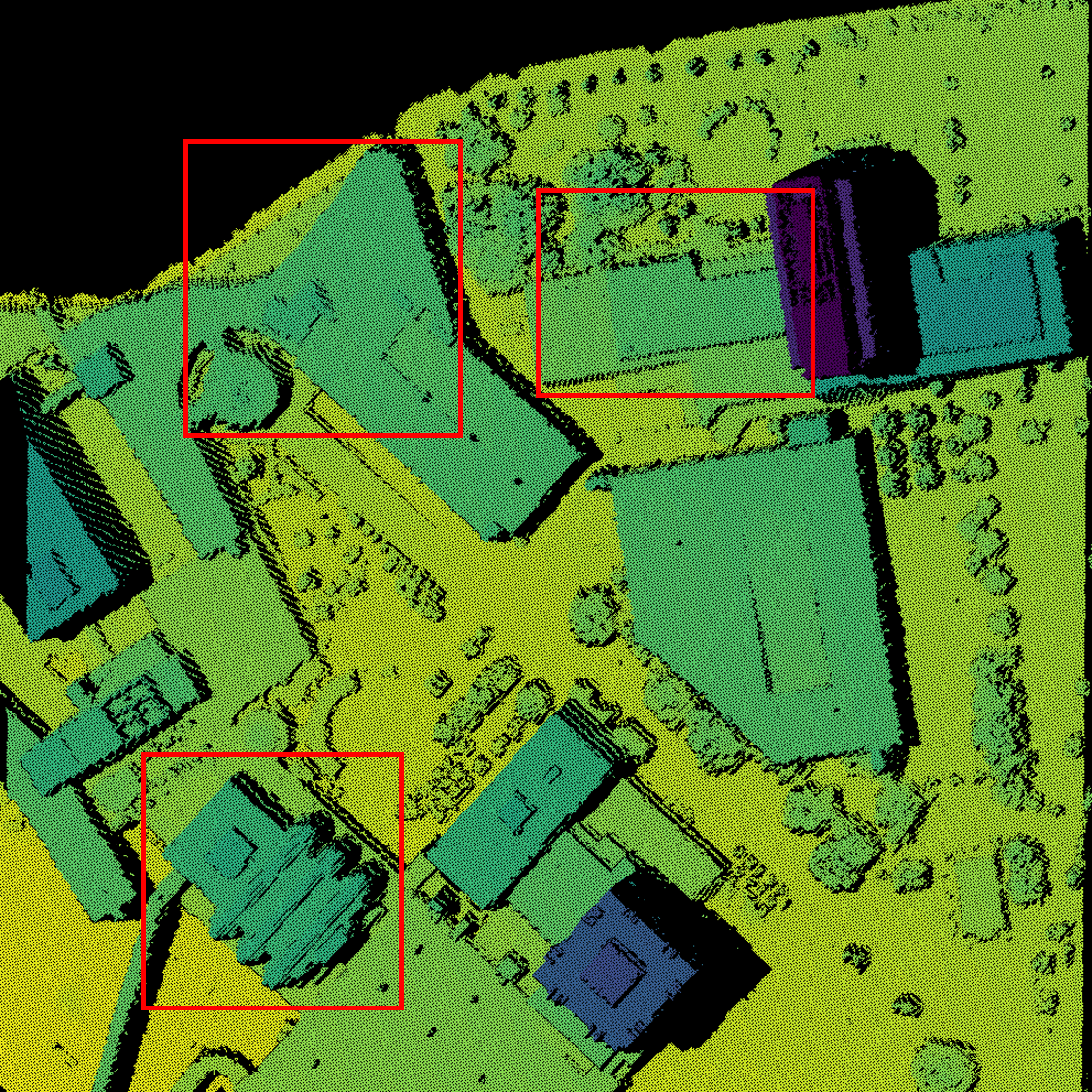}}
  \hfill
  \subfloat[Baseline\label{fig:US3D_ori}]{\includegraphics[width=0.19\textwidth]{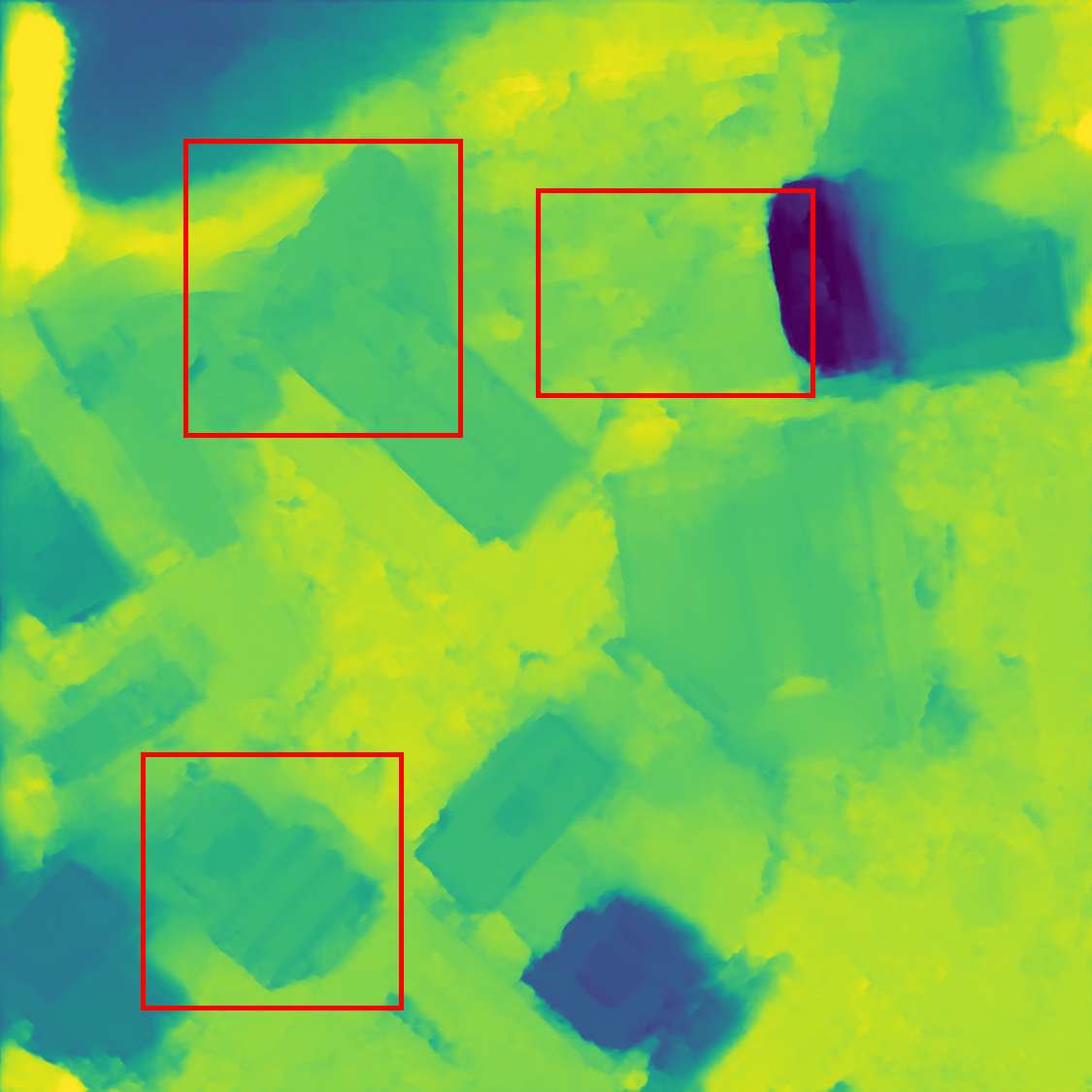}}
  \hfill
  \subfloat[ StereoGAN\label{fig:US3D_stereogan}]{\includegraphics[width=0.19\textwidth]{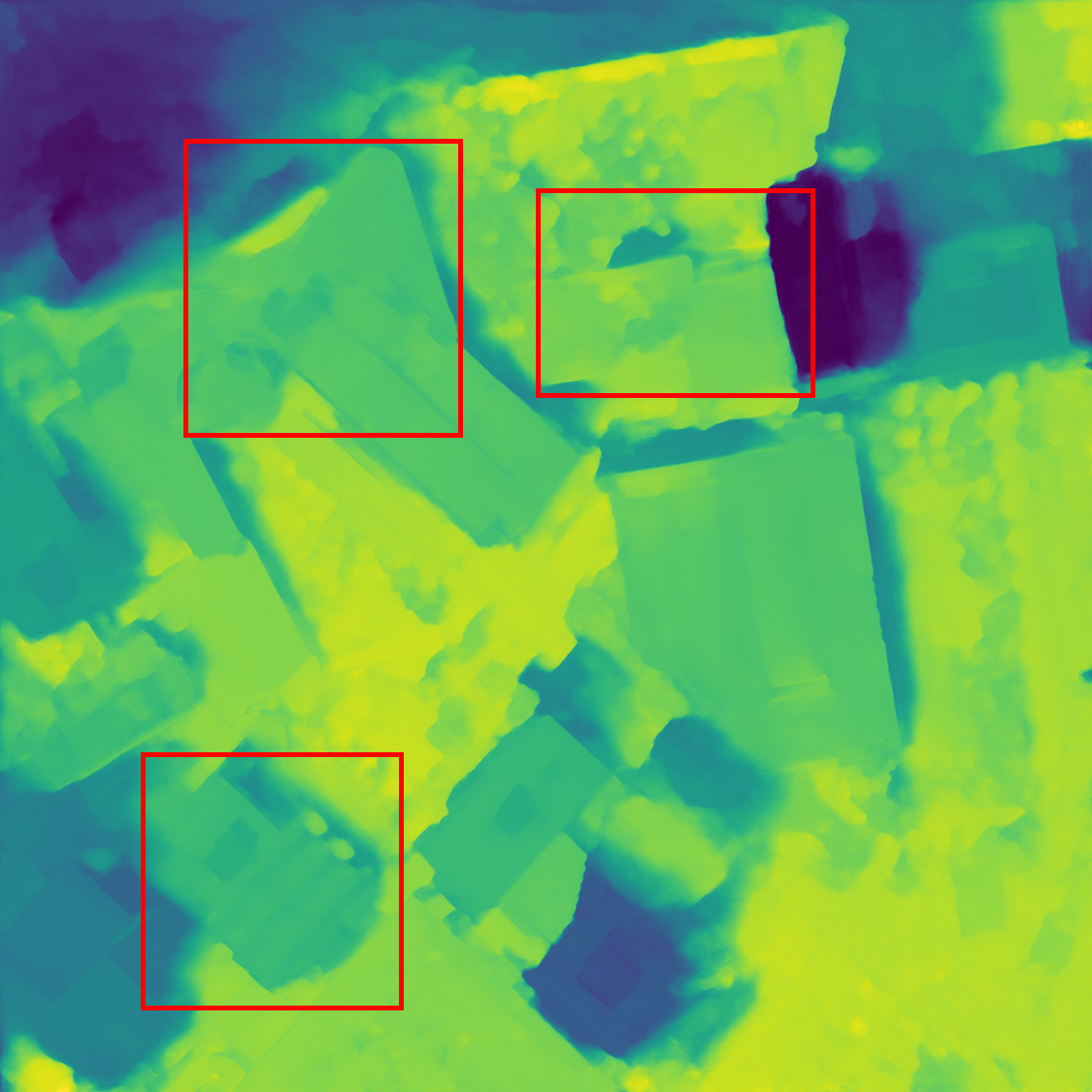}}
  \hfill
  \subfloat[SyntStereo2Real(ours)\label{fig:US3D_secogan}]{\includegraphics[width=0.19\textwidth]{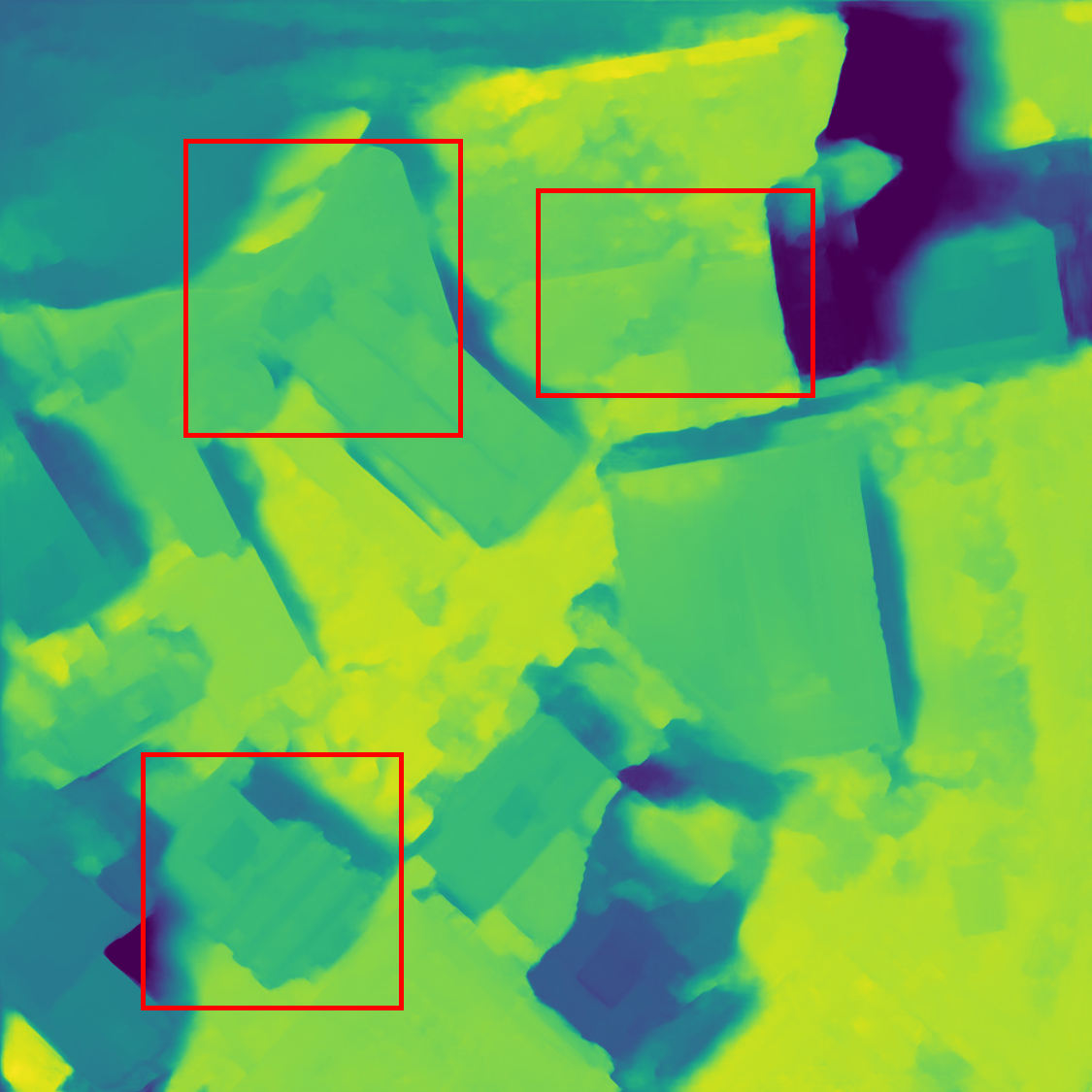}}
   
  \caption{Results of disparity estimation from the AANet for the US3D dataset. Three models are computed for the image shown in (a) RGB reference image, (b) Ground truth, (c) Model trained on SyntCities (Baseline), (d) Model trained on SyntCities translated using StereoGAN (e) Model trained on SyntCities translated using SyntStereo2Real (ours).}
  \label{fig:US3D_disp}
\end{figure*}

\begin{table*}[h!]
\centering
  \begin{tabular}{lcccc}
    \toprule
   \textbf{Metrics} & \textbf{No Edge} & \textbf{With Edge} & \textbf{With Disp} & \textbf{With Edge} \\
     & \textbf{and No Disp} &  &  & \textbf{and Disp} \\
    \midrule
    \textbf{MAD $\downarrow$}& 1.779 & 1.755 & 1.646 & \textbf{1.319} \\
    \textbf{3px-acc\% $\uparrow$}& 62.670 & 62.887 & 63.847 & \textbf{69.906} \\
    \textbf{1px-acc\% $\uparrow$}& 31.503 & 31.853 & 32.600 & \textbf{35.929} \\
    \bottomrule
  \end{tabular}
\caption{Ablation studies. Here the \textit{Edge} refers to the addition of edge information along with input image and \textit{Disp} refers to the additional use of warping loss to enforce disparity constraints.}
\label{tab:ablation}
\end{table*}

\subsection{Network Architecture and Training}
The autoencoder with a pair of encoder and decoder for generator is based on MUNIT
architecture~\cite{munit}. The discriminators are implemented using PatchGAN~\cite{patchgan} architecture. The input to the network consists of images from both the domains and their corresponding
edge maps and the output consists of translated images of both domains with the style and content interchanged.
We recommend the use of Sobel operator to obtain edge maps due to its simplicity and effectiveness, which was compared in~\cite{sda} with other existing edge detectors. The Sobel operator employs two 3 × 3 convolution masks: one for estimating the gradient in the x-direction and the other for the y-direction. This design aligns effectively with CUDA architecture, allowing individual threads to apply the 3 × 3 convolution masks to their assigned pixel and its neighboring pixels in the image.

The network model is implemented using PyTorch~\cite{pytorch} and the training is carried out for $100$ epochs with a batch size of $4$. The hyperparameter values for $\lambda_1$, $\lambda_2$, $\lambda_3$, $\lambda_4$ and $\lambda_5$ in \cref{warp_loss} and \cref{gen_loss} are set to $1$, $1$, $0.8$, $10$ and $10$ respectively. We use stochastic mini batch gradient descent with Adam optimizer~\cite{adam}. Beta coefficients of Adam are set to $0.5$ and $0.999$ respectively.

\subsection{Datasets}

We use two sets of datasets from different application areas to study the generalization capabilities of \textit{SyntStereo2Real} architecture. For remote sensing data, we use SyntCities dataset~\cite{SyntCities} for synthetic data and Urban semantic 3D dataset~\cite{urban3d} for real domain data. SyntCities is a large dataset set consisting of synthetically generated images of aerial imagery. It is specially developed to train deep learning networks for disparity estimation, providing accurate disparity ground truth and different baselines. It consists of 8800 pairs of images resembling architectures of three cities: \textit{New York, Paris and Venice}, with size of each image being $1024\times1024$. We use $1000$ tuples of images taken evenly from all the three cities for training. The Urban3D (US3D) dataset consists of satellite images taken from WorldView3 mission and ground truth disparities are derived from aerial LiDAR data. In this dataset, a significant portion of the images primarily consists of vegetation with limited urban content. To address this, we filtered images based on label data, retaining only those images that contain a minimum of $15$\% building-related content. We randomly selected $1000$ samples each of size $1024\times1024$ for training. 

For autonomous driving data, we use the Driving dataset from SceneFlow~\cite{dispnet} for synthetic domain and KITTI2015~\cite{kitti2015} dataset in real domain. We use the complete dataset from Driving consisting of $4400$ images of size $540 \times 960$ and the $160$ training images each of size $385\times1242$ provided by KITTI2015 benchmark. We resize the images to $512\times512$ for remote sensing dataset and $256\times512$ for autonomous driving dataset during training due to memory and time constraints.

\subsection{Evaluation metrics}

We compare the two models based on performance of stereo matching and number of learnable parameters (compactness of the model architecture) required to train the model.

We acquire translated images and assess their performance on disparity estimation by training them on a disparity network. Specifically, we employ AANet~\cite{aanet} for the training and evaluation of estimation. In case of SyntCities to US3D we trained the model for $400$ epochs and Driving to KITTI 2015 for $120$ epochs, as this is a larger dataset. In both the cases we used a batch size of 20 and the maximum disparity was set to $192$. To evaluate the predicted disparity maps, we removed the areas where the ground truth is not defined. $60$ samples from US3D were used for testing and $40$ for KITTI 2015 (no overlapping with the training samples). The cases where the original data (before translation) is taken as input is named as baseline. 

Given the scarcity of models specializing in synthetic-to-real domain adaptation with stereo constraints, we conduct a comparative analysis of our model against StereoGAN. The results are given in \cref{tab:comparison}. We use MAD (Median Absolute Deviation)~\cite{mad}, $3$px accuracy percentage and 1px accuracy percentage for evaluation of stereo matching. MAD is a robust statistic, being resilient to outliers in a dataset compared to standard deviation because it is calculated by obtaining the median of the absolute difference of pixels and not the squared mean as in standard deviation. $3$px accuracy represents the percentage of pixels in the disparity map for which the estimated disparity is within a range of ±3 pixels from the ground truth disparity and $1$px refers to the same metric but for a 1 pixel range.

\subsection{Quantitative Results}
As indicated in \cref{tab:comparison}, our approach demonstrates enhancements, showcasing a notable improvement with respect to StereoGAN of $+3.14$\% in $3$px accuracy and $+2.30$\% in 1px accuracy for remote sensing images. Besides, the model exhibits improvements of $+1.727$\% in $3$px accuracy and $+1.621$\% in $1$px accuracy for autonomous driving datasets. Please note that the ground truth in the KITTI dataset is sparse and can not be evaluated for all the pixels. Despite that, we can visually compare the reconstruction capabilities for not labelled pixels. The disparity maps illustrated in \cref{fig:KITTI_disp} and \cref{fig:US3D_disp} highlights a more complete prediction without empty regions. Comparing the number of parameters in \cref{tab:num_parameters}, our model has a significantly smaller number of learnable parameters for training, making it ideal for applications with limited storage and processing capabilities.

\subsection{Qualitative Results}
\Cref{fig:results} displays the results of translation of StereoGAN and our network SyntStereo2Real. The main challenge in translating in remote sensing images are maintaining the structural information for all resolution of images. StereoGAN, while proficient in certain aspects of disparity estimation, fails in the translation of shadows by hallucinating green patches instead of building shadows. Our model effectively captures and reproduces the content such as architectural details of building rooftops, bridges and roads, improving shadow handling, and preserving epipolar geometry simultaneously. Our method shows consistent prediction of disparity maps for complete objects without empty gaps or unclear boundaries. We also demonstrate our approach performs well in different application domains beyond autonomous driving. 

\subsection{Ablation Studies}
 In the \cref{tab:ablation}, various configurations of the model are evaluated based on the presence or absence of edge information and warping loss for disparity.  Firstly, the inclusion of edge information results in a decrease in the Mean Absolute Deviation (MAD), indicating improved results in predicting deviations from the ground truth. This decrease, coupled with a corresponding increase in both $3$px accuracy and $1$px accuracy indicates the importance of addition of edge maps. Similarly, addition of warping loss helps in improving the accuracy and MAD of the model significantly.  We also experimented incorporating edges using Spatial feature transform (SFT), where the model learns the weight of edges along with image. This network produced images with strong focus on edges and lost other details making the data unsuitable for training. Thus the the ablation study demonstrates that incorporating both edge information and disparity significantly improves the model's performance across all evaluated metrics for the used datasets.

\section{Conclusion}
In this paper, we propose a light-weight edge based GAN model designed for unpaired image-to-image translation of synthetic-to-real data, while adhering to stereo constraints. Our approach leverages the importance of edge maps along with input images to retain the structural information while translation.  Additionally, we incorporate a warping loss to maintain the accuracy of disparities on translated images. The integration of these two crucial elements  yields state-of-the-art results in a single synthetic to real image translation network.

\section{Acknowledgement}
The authors would like to thank the Johns Hopkins University Applied Physics Laboratory and IARPA for providing the data used in this study, and the IEEE GRSS Image Analysis and Data Fusion Technical Committee for organizing the Data Fusion Contest.

{
    \small
    \bibliographystyle{ieeenat_fullname}
    \bibliography{main}
}

\end{document}